\documentclass[preprint,10pt,authoryear]{elsarticle}

\usepackage{amsmath}
\usepackage{amssymb}
\usepackage{mathabx}
\usepackage{graphicx}
\usepackage{algorithmic}
\usepackage{array}
\usepackage{multirow}
\usepackage{float}
\usepackage{longtable}
\usepackage{hyperref}
\usepackage{cleveref}
\usepackage{graphicx}
\usepackage{caption}
\usepackage{subcaption}
\usepackage{url}
\usepackage{footnote}

\journal{Advanced Engineering Informatics}

\begin{document}
\begin{frontmatter}

\title{Air Traffic Controller Workload Level Prediction using Conformalized Dynamical Graph Learning}

\affiliation[inst1]{organization={Mechanical and Aerospace Engineering},
            addressline={Arizona State University}, 
            city={Tempe},
            postcode={85287}, 
            state={AZ},
            country={USA}}

\affiliation[inst2]{organization={Human Systems Engineering},
            addressline={Arizona State University - Polytechnic}, 
            city={Mesa},
            postcode={85212}, 
            state={AZ},
            country={USA}}

\author[inst1]{Yutian Pang}
\author[inst1]{Jueming Hu}
\author[inst2]{Christopher S. Lieber}
\author[inst2]{Nancy J. Cooke}
\author[inst1]{Yongming Liu\corref{mycorrespondingauthor}}
\cortext[mycorrespondingauthor]{Corresponding author.}
\ead{Yongming.liu@asu.edu}

\begin{highlights}
\item This paper investigates the predictability of air traffic controller (ATCo) workload from spatiotemporal traffic data. 
\item We propose a dynamic graph-learning framework for ATCo workload level prediction with evolving near-terminal traffic patterns.
\item Conformal prediction set is adopted to include uncertainties of predicted workload levels.
\item The proposed study is demonstrated and validated with in-house human-in-the-loop (HITL) simulations with retired ATCos under three real-world scenarios. 
\end{highlights}

\begin{abstract}

Air traffic control (ATC) is a safety-critical service system that demands constant attention from ground air traffic controllers (ATCos) to maintain daily aviation operations. The workload of the ATCos can have negative effects on operational safety and airspace usage. To avoid overloading and ensure an acceptable workload level for the ATCos, it is important to predict the ATCos' workload accurately for mitigation actions. In this paper, we first perform a review of research on ATCo workload, mostly from the air traffic perspective. Then, we briefly introduce the setup of the human-in-the-loop (HITL) simulations with retired ATCos, where the air traffic data and workload labels are obtained. The simulations are conducted under three Phoenix approach scenarios while the human ATCos are requested to self-evaluate their workload ratings (i.e., low-1 to high-7). Preliminary data analysis is conducted. Next, we propose a graph-based deep-learning framework with conformal prediction to identify the ATCo workload levels. The number of aircraft under the controller's control varies both spatially and temporally, resulting in dynamically evolving graphs. The experiment results suggest that (a) besides the traffic density feature, the traffic conflict feature contributes to the workload prediction capabilities (i.e., minimum horizontal/vertical separation distance); (b) directly learning from the spatiotemporal graph layout of airspace with graph neural network can achieve higher prediction accuracy, compare to hand-crafted traffic complexity features; (c) conformal prediction is a valuable tool to further boost model prediction accuracy, resulting a range of predicted workload labels. The code used is available at \href{https://github.com/ymlasu/para-atm-collection/blob/master/air-traffic-prediction/ATC-Workload-Prediction/}{$\mathsf{Link}$}.

\end{abstract}

\begin{keyword}
Air Traffic Management, Aviation Human Factors, Controller Workload, Graph Neural Network
\end{keyword}

\end{frontmatter}


\section{Introduction \label{sec: introduction}}

The rapid advancement of intelligent systems substantially reduces the operational effort from the individual user level but escalates the system-level complexity of real-time decision-making and corporate planning, arises from the dynamically changing environments, time restrictions, and tactical constraints \citep{hancock1988human, sheridan2002humans, nachreiner2006human, loft2007modeling}. Workload assessment and prediction of operating such complex systems have long been regarded as critical research objects \citep{gopher1986workload, gianazza2010forecasting, djokic2010air, tobaruela2014method, wang2015air}. Workload overhead can occur when the demands exceed the human operator's capacity and can lead to efficiency drop and operational safety concerns. Within the aviation domain, effective workload management of air traffic controllers (ATCos) is of utmost importance to maintain safety and rely on accurate ATCo workload predictions.

Air traffic control (ATC) is a crucial part of aviation safety, ensuring that aircraft are safely guided through the airspace and landed or taken off from airports. ATCos are responsible for managing the flow of aircraft, communicating with pilots, and making critical decisions in real-time to ensure the safety of all involved. As air traffic continues to increase \citep{faareport}, it puts more pressure on ATCos, who already have highly demanding and stressful daily routines \citep{ligda2019monitoring, dhief2020predicting, lieber2021deviations}. Quantifying the effort made to meet these task requirements lead to the concept of workload as an air traffic controller, which denotes the subjective qualitative measure of perception demand placed by the current air traffic situation \citep{hilburn2004cognitive, loft2007modeling, djokic2010air}. Moreover, proper workload management and scheduling are vital to ensure ATCos can perform their duties effectively and faultlessly without being overwhelmed. Human performance is a crucial factor in ensuring the safe operations of the National Airspace System (NAS). In the past, human operators have been identified as significant contributors to accidents involving air carrier operations governed by the 14 Code of Federal Regulations (CFR) Part 121, which covers commercial airliners frequently used by the public \citep{regulations2017title}. For instance, approximately 80\% of the 446 air carrier accidents that occurred between 1997 and 2006 were attributed to personnel-related factors, while environmental factors were cited in approximately 40\% of the accidents and aircraft-related factors in 20\% \citep{national2001annual}. Workload prediction can help in several ways, such as ensuring that enough ATCos are available to manage the traffic, preventing fatigue and burnout, optimizing shift schedules, and improving overall efficiency. By accurately predicting the workload, air traffic organizations can ensure that they have the necessary resources and personnel to maintain a safe and efficient air traffic system \citep{pham2020air, heng2022identifying, xiong2023predicting}. All of these objectives are based on reliable ATCo workload-level modeling and predictions. Moreover, artificial intelligence (AI)-enabled human factor studies in aviation have been identified as one of the core elements of AI taxonomy by related authorities \citep{easaAIv1, easaAIv2}, where ATCo workload management is a key dedicated objective.

A tremendous amount of research has been done to understand the impact factors and demand patterns that drive the workload of a controller, such that a better workload prediction performance can be discovered. Two types of factors are studied extensively in the literature, (a) physiological and behavioral features including ATCo mental stress, fatigue level, communication difficulties, and situation awareness \citep{manning2002using, hah2006effect, crutchfield2007predicting, edwards2012factor}; (b) objective factors such as traffic and airspace complexity measures (i.e., operational errors (OE)), abnormal events, level of automation, and weather situations \citep{sridhar1998airspace, chatterji1999neural, chatterji2001measures, majumdar2002factors, edwards2017relationship, corver2016predicting, sharma2022cognitive}. In order to collect features for workload prediction, researchers have proposed to collect human-subject data (i.e., eye movement, communications, heartbeat rates, and Electroencephalography (EEG) signals) \citep{di2010approximation, abbass2014augmented, arico2016adaptive}, in an intrusive and non-intrusive sense. On the other hand, traffic-related features can be directly obtained from computer flight recordings and operational recordings. However, some specific traffic features need \textit{post-hoc} processing (i.e., loss of separations (LoS), OEs). Specifically, in this work, we are interested in traffic-related objective features since it is very unlikely to collect real-time biological features (i.e., EEG/ECG signals or heartbeat rates) in the near future due to privacy concerns and regulatory requirements. Moreover, we discover that the existing model on workload prediction is mostly using handcrafted features, even if a graph data structure and simple neural networks (i.e., minimum spanning trees) have been proposed \citep{chatterji1999neural}. To the best of the authors' knowledge, there is no investigation on utilizing advanced data-driven learning techniques (i.e., graph neural networks (GNNs)) for workload prediction possibilities that directly leverage the spatiotemporal relationships contained in the traffic data and airspace layout. 

In this work, we investigate the possibility of using the graph neural network to predict ATCo workload levels, with an additional post-processing technique, namely conformal prediction, to boost the accuracy with a set of prediction labels. The data is collected by conducting experiments with retired ATC participants who have experience at FAA Radar Approach Control (TRACON) facilities, under three different scenarios, (a) baseline conditions; (b) high workload nominal conditions; (c) high workload off-nominal conditions. The major difference among scenarios are the peak traffic densities and the presence of off-nominal events (i.e., runway switch, communication errors, etc). The simulation scenario is limited to a few Phoenix approach procedures for a duration of 25 minutes for each scenario. A detailed description of the experimental setup is in \Cref{sec: human}. Specifically, the ATCo workload we investigated is the executive (R-side) controllers' workload \citep{pham2020air}. Predicting controller workload levels can be viewed as a pattern recognition problem \citep{chatterji2001measures} and thus is suitable for data-driven learning algorithms. In this work, the problem of predicting workload based on the spatiotemporal layout of airspace is viewed as a time-series dynamically evolving graph classification task. Being time-series classification, we propose to input multiple historical timestamp graphs into the model for the prediction of workload level at the next timestamp. Also, the spatiotemporal layout of the graph structure varies at each timestamp (i.e., number of nodes, graph edge connections), resulting in a dynamical graph classification problem. 

Our contributions are summarized as,
\begin{itemize}
    \item This paper investigates the possibility of predicting executive controller workload during approach scenarios directly from the recorded air traffic data with graph neural networks and discovers that traffic conflict is a nontrivial contributor to improving workload prediction capabilities. 
    \item We propose to formulate the ATCo workload prediction task into a dynamical time-series graph classification problem and show that the Evolving Graph Convolutional Network (EvolveGCN) can achieve a higher prediction accuracy than both statistical (i.e., regression, handcrafted features) and classical learning methods (i.e., MLP, GCN). We show that graph neural networks have great potential for predicting controller workload with varying spatiotemporal airspace layouts. 
    \item A moving window approach is proposed to build the correct input-output matching from the collected sparse workload data. The moving window size represents the temporal length of the historical information used in workload prediction. The selection of parameters can be alternated to fit into the operational need. The data structure formulation transfer complex structured traffic features into a lucid format for research and development purposes.
    \item To further improve the classification accuracy of the experimental data. We explore conformal prediction to expand the prediction as set predictions. We show that conformal prediction has better ground truth label coverage by giving multiple possible predictions as indicators of model uncertainties. We suggest that conformal prediction is a valuable machine learning \textit{post-hoc} processing tool to boost performance further as well as indicate prediction uncertainties.
\end{itemize}

The rest of the paper is organized as follows. First, \Cref{sec: literature review} reviews related studies on air traffic controller workload prediction. We first introduce the impact factors of ATCo workload in \Cref{subsec: workload-factors}, then list the current practices in predicting workload \Cref{subsec: workload-predict}. In \Cref{sec: human}, we introduce the detailed workflow of human-in-the-loop simulations to collect the traffic data and ground truth ATC workload labels, along with data analysis of the collected data. \Cref{sec: methodologies} describes the flowchart of the proposed machine learning framework, from experiment data handling to innovative modeling. The prediction performance and evaluation of the conformal prediction set are discussed in \Cref{sec: experiments}. \Cref{sec: conclusion} concludes this paper by giving limitations of this study and provides future insights.

\section{Related Works \label{sec: literature review}}
 Due to the surging number of daily aviation operations, the aviation industry is in urgent need of advanced decision support tools that can accommodate the rapid annual air traffic growth. Numerous studies have investigated ATCo workload. It's an aviation researchers' consensus that understanding the impact factors that drive mental workload can help improve airspace capacity, thus reducing aviation safety concerns \citep{hilburn2004toward,loft2007modeling,durso2010managing}. With meaningful impact factors collected or modeled, predictive modeling is critical to building an accurate workload prediction algorithm. In this section, we discuss the related works from two aspects, (1) understand the impact factors that drive the mental workload in \Cref{subsec: workload-factors}; (2) discuss the current practice in predicting ATCo workload from open literature with a focus on predicting workload from traffic factors \Cref{subsec: workload-predict}.

\subsection{Task Demands and Impact Factors to ATCo Workload  \label{subsec: workload-factors}}
In air traffic control, task demand refers to the level of mental and physical effort required for ATCos to complete their duties effectively. 
High task demand leads to increased workload, which negatively impacts aviation safety. Therefore, understanding the level of task demand and appropriately managing workload is essential for ensuring that ATCos can perform their duties effectively and safely. Correctly modeling task demand is viewed as the prerequisite for workload prediction for a long history. It's noteworthy to mention that ATCos workload is not a simple function of task demands; the ATC strategy the controller adapted to meet the increased task demands also provides a feedback loop to ATCo workload \citep{loft2007modeling}. We discuss each of these aforementioned grouped impact factors separately.

\textbf{Air traffic factors} refers to both the aircraft count under the ATCo control and their spatiotemporal relationships. The number of aircraft under control is viewed as the most important factor that drives ATCo's mental workload \citep{rose1978air,kopardekar2003measurement}. A high aircraft count leads to higher communication frequency and a higher possibility of safety events, resulting in a higher mental and physical workload. Traffic density is typically measured in aircraft per unit of airspace, such as aircraft in unit time and unit airspace sector area. Measurements of traffic density have been developed based on the averaged vertical/horizontal separation distances \citep{chatterji2001measures}, as they directly infer loss of separations. Other research investigates the necessity to consider flight interactions and flight characteristics, which includes the changes/variability in heading, speed, or altitude, the pattern of how air traffic flows merges and separates into a set of air traffic complexity metrics \citep{delahaye2000air, chatterji2001measures, kopardekar2003measurement, gianazza2006selection, djokic2010air}. Their regression analysis shows subjective workload depends on both aircraft count and other air traffic complexity measures. Additionally, some other studies also suggest that a lower aircraft count also can lead to task overload if these aircraft are interacting in a complex fashion \citep{mogford1995complexity, kallus1999integrated}. 

\textbf{Airspace complexity factors} include the number of routes, altitudes, and restrictions, which can also impact the workload of ATCos, as they need to monitor and manage multiple variables simultaneously. Airspace-related factors are another key contributor to ATCo mental workload \citep{kirwan2001investigating}. Larger airspace size indicates a higher aircraft count and higher metrics on traffic complexity, while small airspace size reduces conflict resolution options and higher traffic evolving rates. It is noteworthy to mention another work considering both the traffic factors and airspace structure complexity and proposes Structural Complexity Metric (SCM), which incorporates a measure of the organization, hierarchy, and interdependence into the complexity calculation \citep{histon2002introducing}. Furthermore, this paper suggests using well-defined ingress and egress points in the airspace to distinguish normal and abnormal flights based on real-time monitoring.  

\textbf{Operational Constraints} are another major contributor that drives the ATCo workload. Operational constraints refer to the temporal variability within the operational conditions of the airspace, as well as the conditions of related technology and equipment. Several factors are viewed as operational constraints; (1) pilot-controller communications are critical for maintaining safe aviation operations. Malfunctions of communication devices can disrupt air traffic control operations. This is known as loss of ratio communication (NORDO) (2) convective weather conditions, such as thunderstorms or heavy fog. These types of objective factors can affect air traffic control operations by reducing visibility and creating unsafe flying conditions. (3) subjective airspace restriction is another type of operation constraint. The restrictions come from multiple sources, i.e., aircraft holding, no-fly zones, or special-use airspace \citep{loft2007modeling}. In addition, certain other off-nominal events are considered operational constraints, i.e, runway switch, and minimum fuel reported \citep{wickens2009identifying, fraccone2011novel}. 

\textbf{Cognitive states} directly contribute to the cognitive task demands of ATCo. To measure cognitive states, the researchers propose to measure the physiological states of the air traffic controller, including brain activities, eye movements, and heartbeat rates. These states can be quantitatively measured by sensors signals such as electrocardiography (ECG) signals, electroencephalography (EEG) signals, galvanic skin response (GSR), blood pressure (BP), and certain biochemical analysis \citep{crump1979review, vogt2006impact, trapsilawati2020eeg}. However, using intrusive physiological state measurements is disruptive to controllers' normal working conditions, as it creates additional mental stress and discomfort in maintaining ATC operations. Alternatively, computer vision (CV) based non-intrusively physiological state measurements are proposed to collect distractions, drowsiness, head poses, eye movements, and fatigue levels \citep{li2019hybrid, wee2019integrated, xiong2023predicting}. However, these types of measurements can lead to information security and privacy concerns \citep{liang2020behavioral, berghoff2021interplay}.

\subsection{Workload Prediction Algorithms \label{subsec: workload-predict}}

The dynamic density model builds a regression model to find the linear relationships between traffic complexity factors and ATCo workload. The Dynamic Density metric uses a combination of traffic density and complexity measures to estimate task demands in real-time, with the goal of providing a more accurate and responsive measure of controller workload from task demands \citep{masalonis2003dynamic}. However, dynamic density metrics fail to consider human cognitive capacities, which are the primary source of ATCo workload sources in the real world. In \cite{kopardekar2003measurement}, the traffic complexity, as well as the airspace complexity of different sections, are considered. The results show that the airspace factor can actually contribute to workload prediction in a multi-sector study. Similarly, in \cite{hah2006effect}, the authors find that the ATCo workload is proportional to the number of aircraft controlled by the enroute sector. They conduct HITL experiments and found a linear relationship between aircraft count and workload ratings. 

However, it's still difficult to identify the most contributing impact factors to workload prediction from regression analysis. The first reason is the multi-collinearity within these factors. For instance, the number of conflicts depends on the speed, altitude, and heading variabilities. The complexity of traffic situations, such as traffic density and potential conflicts, also mediate the causal connection between traffic count and workload. The inter-relationship of these factors makes it challenging to determine the relative importance of each predictor in a regression equation. The second query is the debate on the linear relationship between these factors and workload ratings. For instance, the ATCo can alternate control strategies during a certain period to meet the increased task demands. Additionally, the online processing or post-processing of these factors only considers the current situations, and there is no inference on ATCo's intent and air traffic intent information. Trajectory prediction from flight plans helps with estimating the workload of ATCos, where prediction and reduction of trajectory uncertainties can help alleviate ATCo workload \citep{knorr2011trajectory, corver2016predicting}.

Machine learning algorithms have been adopted for workload prediction. In \cite{gianazza2017learning}, tree-based models, and support vector machines are included for workload measurements. The authors consider both traffic complexity and operational constraints as features and show a high F1 score of over 0.9. However, these types of works are not dynamically considering the \textit{traffic pattern} of the airspace, but still formulate the data as a tabular format, which fails to address the aforementioned concerns \citep{loft2007modeling}. On the other hand, \cite{wang2015air} proposes to use a 3-layer simple neural network to forecast ATCo workload. However, the ATCo workload in this work is assessed from voice communication data, which fails to model the task demand factors mentioned earlier. Impressively, direct prediction from the spatiotemporal air traffic layout is actually proposed decades ago. In \cite{sridhar1998airspace, chatterji2001measures}, the authors propose to model air traffic at each timestamp into graph-structured data and calculate the second-order statistics of time-series of graphs as extracted air traffic complexity measurements. Then a simple neural network is adopted to do classification from these features.

In summary, understanding the factors that drive ATCo workload has been a challenging yet unresolved open question for decades. Instead of investigating the linear relationship between impact factors and workload, one should look into the dynamic properties of these factors and workload. This leads to our study on workload prediction -- we model the spatiotemporal airspace layout into time-series graph structures and propose to use a time-series learning algorithm to predict workload levels, in consideration of historical dynamic variabilities contained within the air traffic data.

\section{Human-In-The-Loop (HITL) Simulations \label{sec: human}}
In this section, we provide an overview and the detailed simulation setup of the Human-In-The-Loop (HITL) experiment as in \Cref{fig: hitlsetup}. The scope and objectives will be discussed. Then, data analysis on the collected data is presented in \Cref{subsec: hitl-analysis}.

\subsection{Simulation Overview}
The HITL simulation is the first human factor study of our aviation big data project, which aims at addressing the safety needs and technology solutions for future NAS \citep{liu2018information}. The backbone of the project lies in information fusion and uncertainty management for real-time system-wide safety assurance, where human factors like ATCo workload play a key role. As mentioned, accurately predicting the ATCos workload can improve operational efficiency and reduce safety concerns such that aviation authorities can ensure a reasonable resource allocation and workload management.

\begin{figure}
    \centering
    \includegraphics[width=\textwidth]{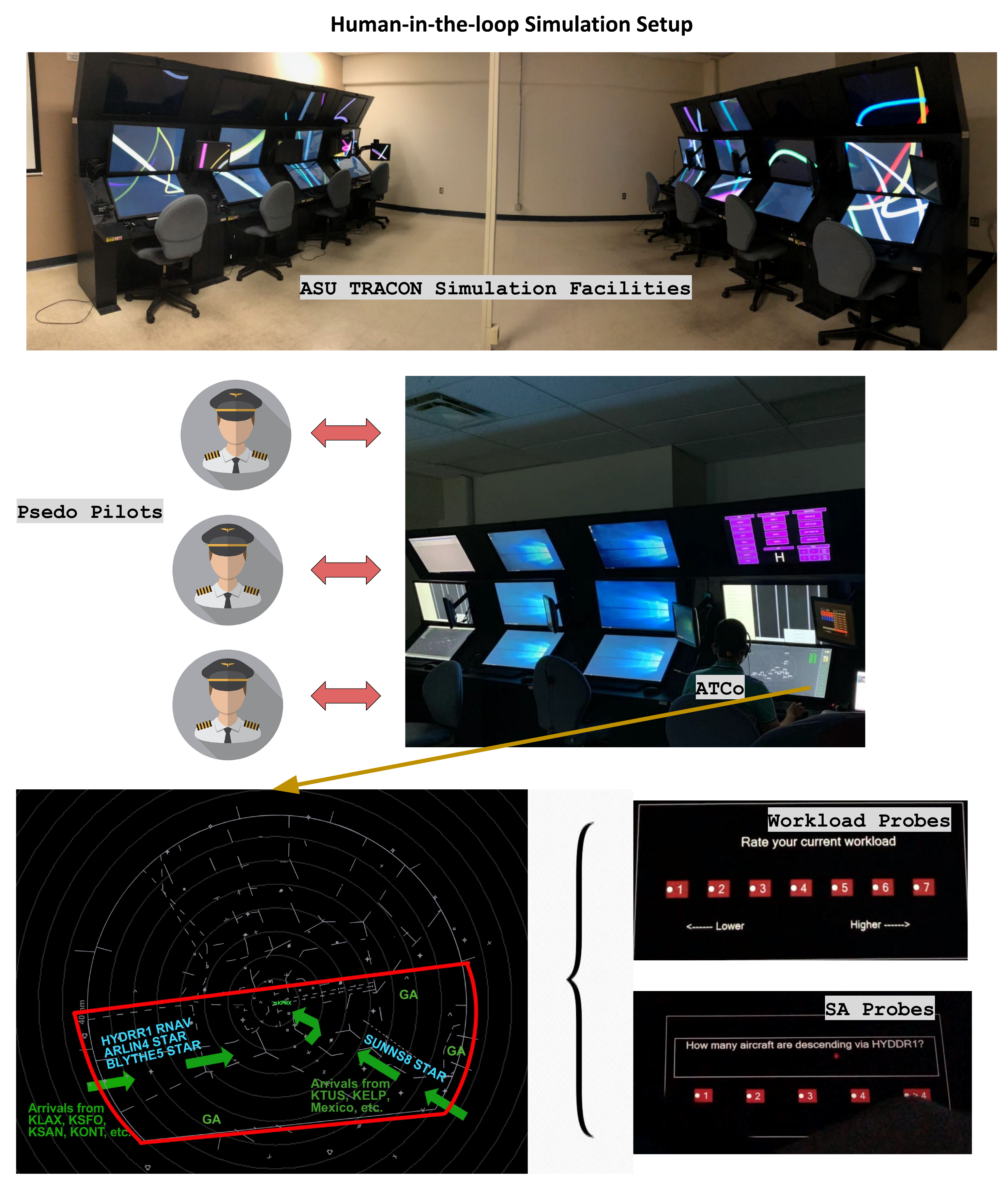}
    \caption{Human-In-The-Loop (HITL) experiment setup. The ASU TRACON Simulation facility is equipped with eight simulators. During the HITL experiments, three pseudo pilots act as pilots and interact with ATCo. We show a simple demonstration of the graphical user interface, where the primary focus of the simulation is on the KPHX arrivals from two directions, with flight procedures including 1 RNAV (HYDRR1) and 3 STARs (ARLIN4, BLYTHE5, SUBSS8). During the experiment, the ATCo is asked to respond to the questions shown on the pop-up window. The window acts as either a workload probe or a situational awareness probe, showing every 3 minutes. }
    \label{fig: hitlsetup}
\end{figure}

\begin{figure}
    \centering
    \includegraphics[width=\textwidth]{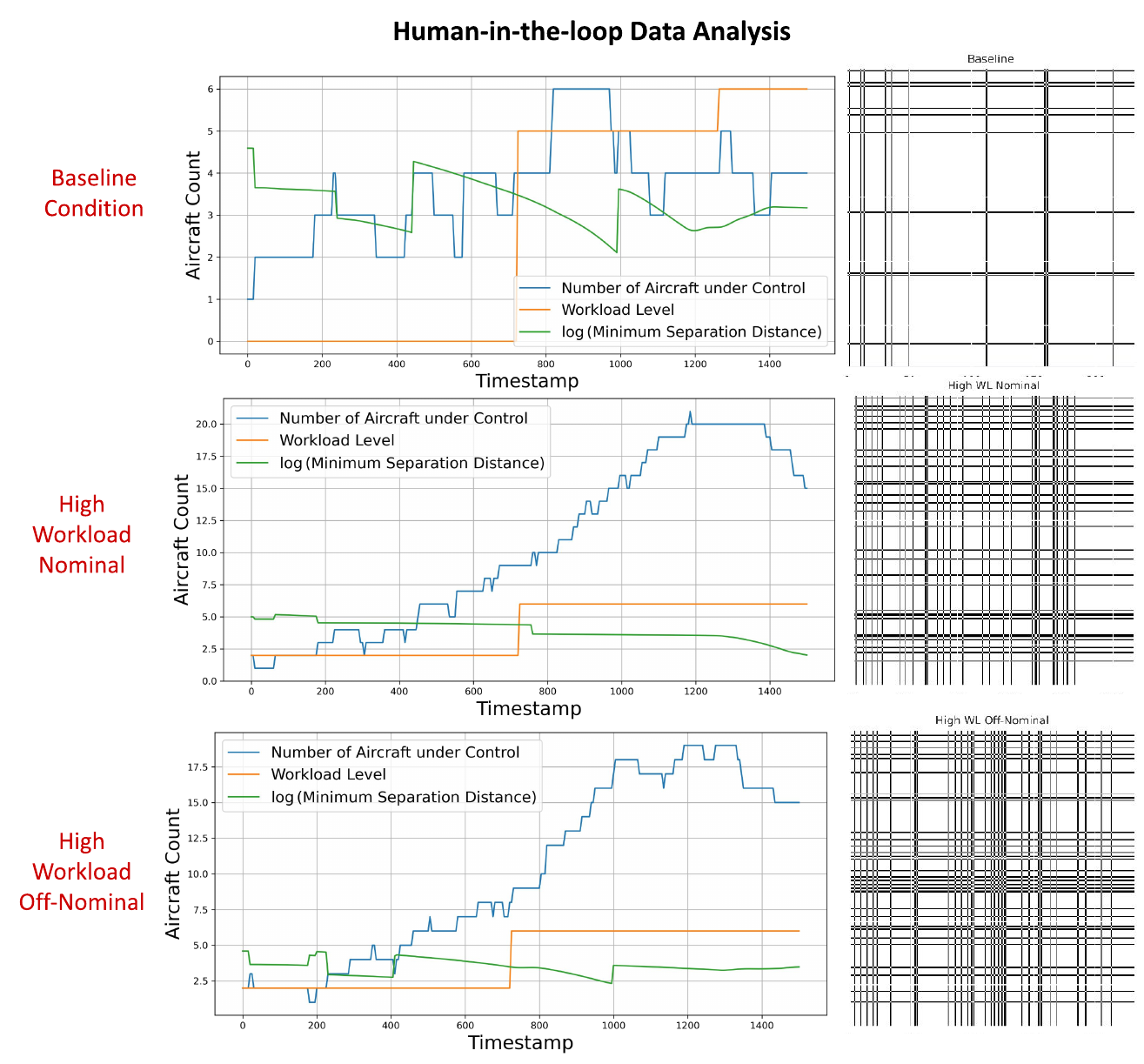}
    \caption{The visualization of collected data samples from the HITL experiments under three scenarios. Each ATCo was involved in three experiments under normal, high workload nominal. high workload off-nominal working scenarios, with a time duration of 25 minutes each. Traffic-related features (e.g., aircraft numbers, minimum separation distance) are collected every 5 seconds. Workload levels are interpolated to match the corresponding traffic-related features at each timestamp. Additionally, we obtain the recurrence plot (RP) from communication transcripts as indicators of system tendency in \Cref{fig: hitl-eval-rrmap}.} 
    \label{fig: collecteddata}
\end{figure}

\begin{table}[htb]
\caption{A short view of the communication transcripts post-processed from radio recordings in HITL simulation. Three cut-off sections are listed here, which correspond to off-nominal events, (1) turbulence reported; (2) no radio communications; (3) landing runway switch. Indicator of communication deviations is also shown in the right-most column. A mark of "$0$" if the ATCo communication followed a pilot's communication or vice versa, and "$1$" will be noted if the communications are not effective and are immediately followed between the ATCo and the pilot.}
\label{table: communications}
\resizebox{\textwidth}{!}{
\begin{tabular}{lllll}
\hline
\multicolumn{1}{l|}{Speaker}                                                                  & \multicolumn{1}{l|}{\begin{tabular}[c]{@{}l@{}}Start \\ Time\end{tabular}} & \multicolumn{1}{l|}{Transcriptions}                                                                                                                                                  & \multicolumn{1}{l|}{\begin{tabular}[c]{@{}l@{}}End \\ Time\end{tabular}} & \begin{tabular}[c]{@{}l@{}}Deviation\\ Indicators\end{tabular} \\ \hline
\multicolumn{5}{c}{... ...}                                                                                                                                                                                                                                                                                                                                                                                                                                                                                   \\ \hline
\multicolumn{1}{l|}{Speed bird 281}                                                           & \multicolumn{1}{l|}{05:26.9}                                               & \multicolumn{1}{l|}{\begin{tabular}[c]{@{}l@{}}apporach speed bird two eighty-one we are experiencing \\ motor turbulence at one three thousand\end{tabular}}                        & \multicolumn{1}{l|}{05:30.5}                                             & 1                                                              \\ \hline
\multicolumn{1}{l|}{\begin{tabular}[c]{@{}l@{}}PHX approach \\ (Speed bird 281)\end{tabular}} & \multicolumn{1}{l|}{05:34.9}                                               & \multicolumn{1}{l|}{speed bird two eighty-one heavy, roger}                                                                                                                          & \multicolumn{1}{l|}{05:37.7}                                             & 1                                                              \\ \hline
\multicolumn{5}{c}{... ...}                                                                                                                                                                                                                                                                                                                                                                                                                                                                                   \\ \hline
\multicolumn{1}{l|}{\begin{tabular}[c]{@{}l@{}}PHX approach \\ (Cumpacity 250)\end{tabular}}  & \multicolumn{1}{l|}{11:32.8}                                               & \multicolumn{1}{l|}{\begin{tabular}[c]{@{}l@{}}cumpacity two fifty descending maintain\\ four thousand one hundred\end{tabular}}                                                     & \multicolumn{1}{l|}{11:35.4}                                             & 0                                                              \\ \hline
\multicolumn{1}{l|}{\begin{tabular}[c]{@{}l@{}}PHX approach \\ (Cumpacity 250)\end{tabular}}  & \multicolumn{1}{l|}{11:38.8}                                               & \multicolumn{1}{l|}{cumpacity two fifty four thousand one hundred}                                                                                                                   & \multicolumn{1}{l|}{11:40.5}                                             & 1                                                              \\ \hline
\multicolumn{1}{l|}{Cumpacity 250}                                                            & \multicolumn{1}{l|}{11:41.8}                                               & \multicolumn{1}{l|}{cumpacity two fifty dropping down to four one hundred}                                                                                                           & \multicolumn{1}{l|}{11:44.6}                                             & 1                                                              \\ \hline
\multicolumn{5}{c}{... ...}                                                                                                                                                                                                                                                                                                                                                                                                                                                                                   \\ \hline
\multicolumn{1}{l|}{Local south}                                                              & \multicolumn{1}{l|}{14:29.1}                                               & \multicolumn{1}{l|}{\begin{tabular}[c]{@{}l@{}}court this is local south we are switching to runway seven \\ left and right effect immediately\end{tabular}}                         & \multicolumn{1}{l|}{14:35.9}                                             & 1                                                              \\ \hline
\multicolumn{1}{l|}{\begin{tabular}[c]{@{}l@{}}PHX approach\\  (Local south )\end{tabular}}   & \multicolumn{1}{l|}{14:38.8}                                               & \multicolumn{1}{l|}{Ok…move to runway seven}                                                                                                                                         & \multicolumn{1}{l|}{14:40.7}                                             & 1                                                              \\ \hline
\multicolumn{1}{l|}{\begin{tabular}[c]{@{}l@{}}PHX approach \\ (Ascer 4527)\end{tabular}}     & \multicolumn{1}{l|}{14:49.8}                                               & \multicolumn{1}{l|}{\begin{tabular}[c]{@{}l@{}}shuttle forty-five twenty-seven expect dail is runway seven \\ left turn left in two seven zero maintain  five thousand\end{tabular}} & \multicolumn{1}{l|}{14:55.1}                                             & 1                                                              \\ \hline
\multicolumn{5}{c}{... ...}                                                                                                                                                                                                                                                                                                                                                                                                                                                                                   \\ \hline
\end{tabular}
}
\end{table}

The primary objective of the HITL experiment is to investigate the correlation between communication patterns (such as content, volume, and flow patterns) and both controller workload and human performance. \Cref{fig: hitlsetup} gives the overview setup of the HITL simulation process. The simulation contains two arrival flows including four Phoenix in-bound Procedures. The top panel of \Cref{fig: hitlsetup} displays the layout of the ASU TRACON Simulation facility. The ASU TRACON Simulation facility boasts a total of eight advanced simulators, designed to provide an immersive training experience. Within the context of HITL (Human-in-the-Loop) experiments, the simulation involves the participation of three pseudo-pilots who assume the role of actual pilots and engage in interactive communication with Air Traffic Controllers (ATCo).

To illustrate the capabilities of the simulation, we present a captivating demonstration of the graphical user interface. The simulation places particular emphasis on the KPHX (Phoenix Sky Harbor International Airport) arrivals from two distinct directions. The first arrival flow represents flights from the west coast (KLAX, KSFO, KSAN, KONT, etc), with procedures HYDRR1, ARLIN4, and BLYTHE5. The second flow stands for arrival flights from the southeast, including KTUS, KELP, and Mexico. SUNSS8 arrival procedure is used here. These procedures challenge both pilots and ATCos to execute precise maneuvers and coordinate their actions effectively. 

Throughout the experiment, the ATCo is presented with a series of thought-provoking questions displayed in a pop-up window. This window either acts as both a workload probe to test the ATCo's ability to manage multiple tasks or a situational awareness probe, evaluating their understanding of the ongoing situation. The questions are presented at regular intervals of three minutes, ensuring a continuous evaluation of the ATCo's performance and cognitive response. Each retired ATCo participant engaged in a within-subjects (3 simulation trials) study design. Each trial spanned 25 minutes and varies in workload level by manipulating two variables, namely traffic density, and occurrence of off-nominal events.

\textbf{Baseline}: Baseline trials contain up to 6 aircraft in the airspace at a given timestamp. There are no off-nominal events in baseline trials. Typically, a moderate workload is expected.

\textbf{High Workload Nominal}: High workload nominal trials can have up to 21 aircraft showing up in the current simulation environment. Again, there are no off-nominal events.

\textbf{High Workload Off-Nominal}: In addition to the experimental setup in high workload nominal trials, high workload off-nominal trials incorporate four off-nominal events during the 25 min duration. We list the name of these off-nominal events here,
\begin{itemize}
    \item Turbulence: Moderate turbulence is simulated in several arrival flows. In \Cref{table: communications}, speed bird 281 reported experiencing turbulence at a certain altitude, starting from 05:26.9 of simulation time.
    \item No radio (NORDO): The pilot has no radio communication with the approach ATCo, which can happen during a radio failure. In \Cref{table: communications}, the KPHX approach controller repeated the order when the first order at 11:32.8 was not confirmed. 
    \item Runway switch: In this simulation, the landing runway switch from KPHX 25L to 07R. The order is given by the local tower, as in \Cref{table: communications} 14:29.1. 
    \item Minimum fuel: At the end of each simulation trial, the aircraft encountered fuel issues.
\end{itemize}
These corresponding timestamps $t_1$, $t_2$, $t_3$, and $t_4$ to apply these off-nominal events is indicated in \Cref{fig: hitl-eval-rrmap}, respectively. 

\subsection{Simulation Setup}
HITL was conducted in eight air traffic management system Metacraft facilities located at the Arizona State University TRACON Simulation Lab, which can be operated as either ATC terminal radar positions or pseudo-pilot stations. The human controllers were retired ATCos who have experience with civilian TRACON facilities within the past 15 years but do not persist possess experience with Phoenix TRACON \citep{ligda2019monitoring}. Six retired ATCos were involved in this study. There were also three researchers who act as pseudo-pilots to fly along the assigned arrival routes during each simulation scenario. Metacraft is the name of the TRACON radar simulation computer cluster system. It provides ATC functions to maneuver the aircraft in a simulation environment, including altitude, speed, and heading. Metacraft collects and maintains data logs such as spatiotemporal tracks, LoS events, and distance measures. 

During each 25 minutes experiment trial, the pop-up window showed a questionnaire probe every 3 minutes, asking either a question on workload rating or situational awareness questions. Specifically, the workload rating questions were shown three times at exact 3 min, 12 min, and 21 min timestamps. \Cref{fig: collecteddata} visualizes the collected data for one participant of three scenarios. Features include minimum separations (traffic conflict), number of aircraft (traffic density), and workload ratings are reported. A recurrence plot indicating communication tendency is also provided, visually representing the communication efforts between the tower controller and the pseudo-pilots. 

The workload rating probe was designed based on the subjective workload assessment technique (SWAT) \citep{reid1988subjective}. The SWAT method is a situation-present assessment method that is commonly used in human factors research. It involves participants rating their perceived workload using various rating scales, such as the NASA task load index (TLX) scale \citep{hart1988development, hart2006nasa}. The SWAT method also includes measures of mental effort and task difficulty to provide a more comprehensive assessment of workload. The workload probe employs a two-step process for administering questions related to situation awareness or workload. Participants first press a ready button, followed by selecting a response. The timing of both actions is recorded, following the methods used in the aforementioned studies. Another important measure of SWAT was the behavioral measure of workload, the time to respond to the ready button.

The controller workload is self-evaluated by the workload question pop-up window, and the human performance is indicated by the count of separation violations. To facilitate this investigation, we have gathered preliminary data on three types of metrics: 1) aeronautical separation violations, which are viewed as traffic conflicts existed in the airspace and an indicator of ATCo performances; 2) real-time workload ratings, taken at three different points in each 25-minute scenario; and 3) audio recordings of controller-pilot transmissions during the workload ratings. These initial settings will serve as a foundation for further analyses using additional measures, such as facial recognition, heart rate variability, situation awareness probes, and operational efficiency. Ultimately, this research provides a solution for the development of a real-time controller workload level prediction system. 

\begin{table}[H]
\caption{Description of \textit{selected} features recorded in the HITL simulations. Traffic density is directly obtained from the Metacraft. The latency variables are defined and collected following modified SWAT. Workload ratings are collected from the question probe. Latency measures and situational awareness questions are collected but not used in this work.}
\label{table: feature_descriptions}
\resizebox{\textwidth}{!}{
\begin{tabular}{c|c|c}
\hline
\textbf{\begin{tabular}[c]{@{}c@{}}Feature \\ Names\end{tabular}} & \textbf{\begin{tabular}[c]{@{}c@{}}Feature \\ Descriptions\end{tabular}}                                                                           & \textbf{\begin{tabular}[c]{@{}c@{}}Feature \\ Values\end{tabular}} \\ \hline
\textit{traffic\_density}                                         & Total number of aircraft under ATC participant's control                                                                                           & Integer: 0-23                                                      \\ \hline
\textit{ready\_latency}                                           & \begin{tabular}[c]{@{}c@{}}Time spent from screen appearing "Ready?" to \\ participant pressing "Ready?" on the pop-up questionnaire.\end{tabular} & Decimal: 0.01-60.00 sec                                            \\ \hline
\textit{query\_latency}                                           & \begin{tabular}[c]{@{}c@{}}Time spent from pressing "Ready?" to selecting answers \\ on the pop-up questionnaire\end{tabular}                      & Decimal: 0.01-60.00sec                                             \\ \hline
\textit{wl\_rating}                                               & Workload rating select from the pop-up window                                                                                                      & Integer: 1-7                                                       \\ \hline
\textit{sa\_correct}                                              & \begin{tabular}[c]{@{}c@{}}Evaluation of selected responses on situation awareness questions\\  from the pop-up window\end{tabular}                & 0: no resp; 1: correct; 2: incorrect                               \\ \hline
\end{tabular}
}
\end{table}

In this paper, we obtain the flight traffic recordings and the real-time workload ratings from real-world human-in-the-loop simulations. The subjective workload rating is collected from the question pop-up windows showing at 3 min, 12 min, and 21 min for each trial. The originally collected data and adjusted workload rating score has been discussed in the literature \citep{lieber2020communications}. By doing this, we obtain realistic human workload levels, or the ground truth, from participants' honest ratings of their mental status. This is the most reasonable data source for obtaining features and labels for building real-world machine-learning pipelines. 

\subsection{Empirical Data Analysis \label{subsec: hitl-analysis}}
Communication transcription analysis is performed based on post-processed radio recordings. There are three major components in this part, (a) use a speech recognition tool or manual transcription tool to translate voice to text; (b) identify the named entity of each communication transcript (i.e., controllers or pilots); (c) perform either statistical analysis or keyword extraction \citep{gorman2003evaluation, salas2004team, cooke2017communication}. As mentioned, the deviation indicator represents the deviation in communications, also known as closed loop communication deviation (CLCD) \citep{lieber2021deviations}. CLCD is based on an established coding scheme derived from the expected exchange of closed loop communication (CLC). Deviations occur when consecutive pilot communications or consecutive air traffic controller communications take place. CLCs were coded using a binary system to detect CLCD based on communication patterns between pilots and ATC. An expected CLC pattern involves alternating communications between pilots and ATC. CLCD is identified when a pilot's communication follows another pilot's or when consecutive ATC communications occur.

\begin{figure}[H]
    \centering
    \includegraphics[width=\textwidth]{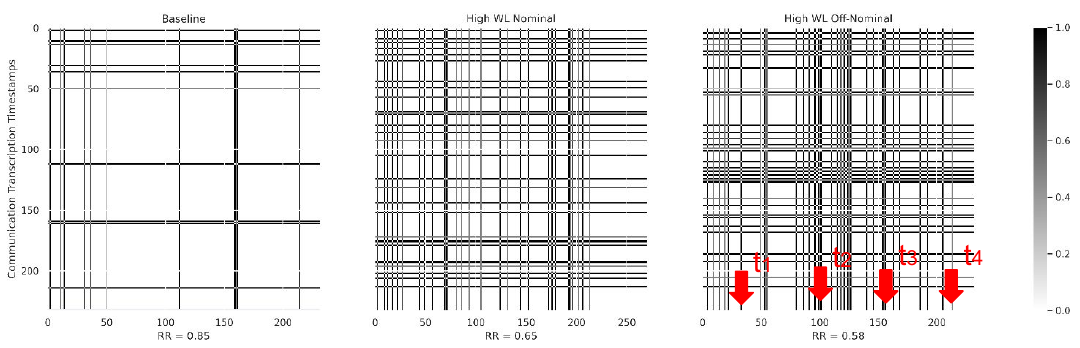}
    \caption{Communication Transcription Visual Analysis: Recurrence Plot (RP). RP is used to quantify the overall tendency of recurrence in the system. Vertical/Horizontal lines indicate the laminar states do not change or change slowly over time \citep{MARWAN2007237}.}
    \label{fig: hitl-eval-rrmap}
\end{figure}

\Cref{fig: hitl-eval-rrmap} shows the closed loop communication deviation analysis's recurrence plot (RP). RP was originally proposed to visualize the complexity of dynamical systems \citep{eckmann1995recurrence, marwan2007recurrence}, where a detailed mathematical formulation can be found. In this work, we define the phase vector as the communication deviations and build the recurrence matrix $\mathbf{R}$ as,

\begin{equation}
    \mathbf{R}_{i,j} = \left\{ \begin{array}{rcl}
1, & \mbox{for} & \vec{x_i} \approx \vec{x_j} \\ 
0, & \mbox{for} & \vec{x_i} \not\approx \vec{x_j} \\
\end{array}\right. i,j = 1, ..., N
\end{equation}

where N indicates the number of current states. $\vec{x_i} \approx \vec{x_j}$ means that they are approximately equal up to an error round defined as $\varrho$. In general, the recurrent matrix $\mathbf{R}$ compares the state and indicates the state similarity across the entire series \citep{marwan2007recurrence}. The selection of similarity threshold $\varrho$ is critical. Researchers have investigated the selection criterion in the literature based on the system states \citep{koebbe1992use, zbilut1992embeddings, mindlin1992topological}. In \Cref{fig: hitl-eval-rrmap}, we choose $\varrho=0.1$ for the visualization. Moreover, \Cref{fig: hitl-eval-rrmap} suggest a typical vertical and horizontal lines pattern, which is suggested to be a laminar state or state idle case, while the sparse region indicates lower system complexity and vice versa \citep{marwan2007recurrence}.

\begin{figure}
    \centering
    \begin{subfigure}[t]{0.45\textwidth}
        \centering
        \includegraphics[width=\textwidth]{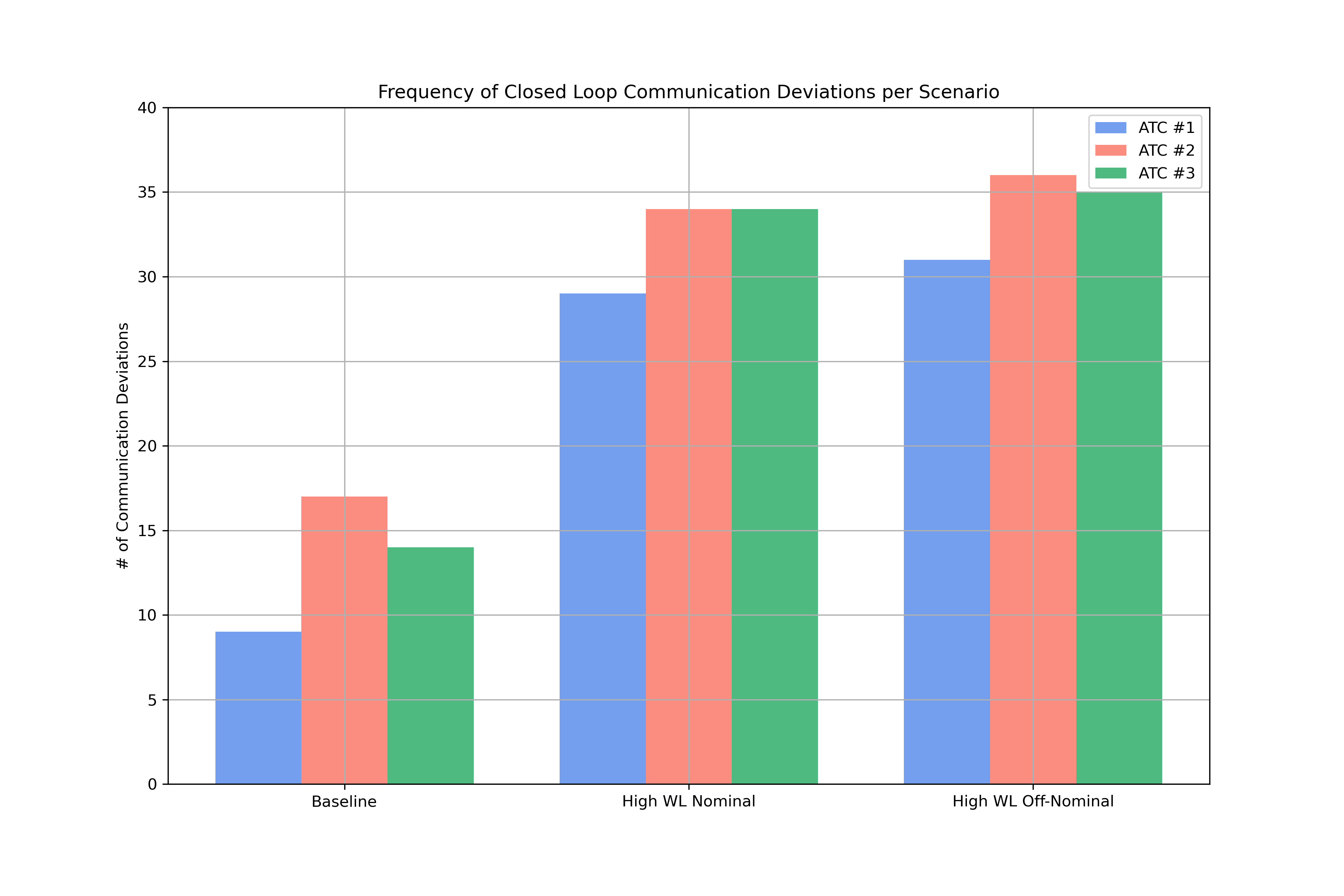}
        \caption{Number of Deviations that happened in each scenario. Showing first three ATCs.}
        \label{fig: hitl-eval-freqdev} 
    \end{subfigure}
    ~
    \begin{subfigure}[t]{0.45\textwidth}
        \centering
        \includegraphics[width=\textwidth]{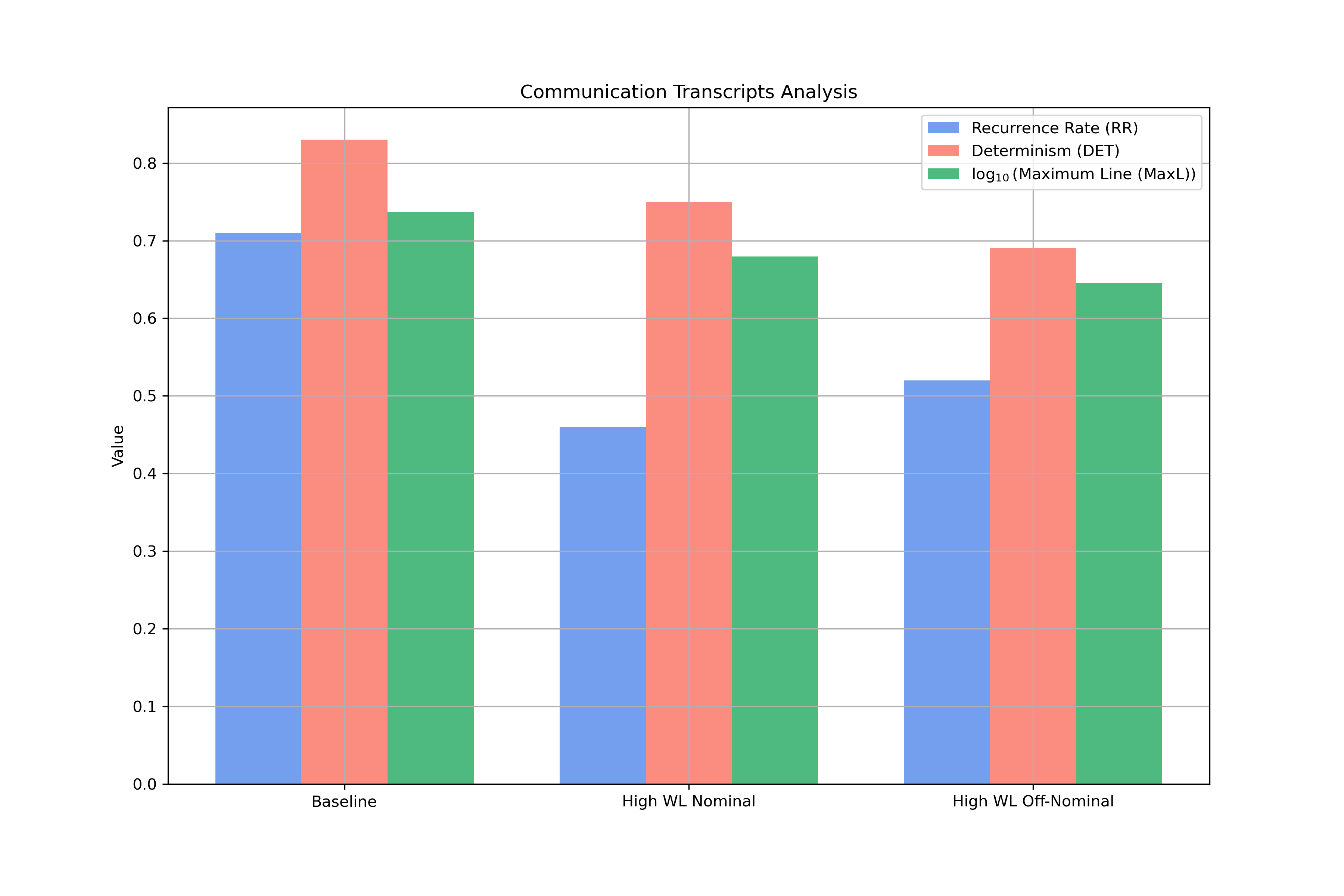}
        \caption{Recurrence Quantification Analysis (RQA).}
        \label{fig: hitl-eval-rqa} 
    \end{subfigure}
    \caption{Communication data analysis on different scenarios. In (a), we show the histogram of the frequency of the communication deviations for each scenario (up to 3 ATCos). As discussed, the number of communication deviations indicates communication difficulties. In (b), we show the QRA under three different scenarios, showing the scenario complexities obtained from communications.}
    \label{fig: hitl-eval}
\end{figure}

Beyond the visual approach, recurrence quantification analysis (RQA) is also widely used to measure system-level complexity \citep{zbilut1992embeddings, webber1994dynamical, marwan2002recurrence}. Typically, RQA is based on the diagonal and vertical patterns of the RP. We use three types of complexity measures here.

\textbf{Recurrence Rate (RR)} is the simplest measure of RP, which is the averaged density of recurrence points in RP. RR represents the likelihood of a state returning to its $\varrho$-neighborhood in the phase space \citep{kantz1994quantifying}. It's the measure of correlations between the ATCo and pilot communications. 

\textbf{Determinism (DET)} refers to the degree of predictability or orderliness in a system's dynamics over time. A deterministic system is one in which future states can be precisely predicted from knowledge of the present state and the system's dynamics. In the context of recurrence plots, a high degree of DET is indicated by the presence of diagonal lines in the plot, which represent points in the system's trajectory that are close to each other in phase space and recur with a high degree of regularity. Conversely, a low degree of DET is indicated by a more random or chaotic pattern in the recurrence plot, with fewer or no diagonal lines. It indicates the predictability of ATCo and pilot interaction.

\textbf{Maximum Line (MaxL)} is a diagonal line that represents the longest connected sequence of recurrent points in the plot. It is the diagonal line that has the most points along it, and it indicates the most persistent pattern of recurrence in the system's dynamics. The length and frequency of maximum lines can provide insights into the regularity and predictability of the system's behavior over time. MaxL quantifies the stability of ATCo and pilot interaction.

\Cref{fig: hitl-eval} shows the histograms of communication deviations in (a), and the calculated measures of complexity in (b). As shown in Figure \ref{fig: hitl-eval}(a), there is a notable rise in communication challenges from the baseline scenarios to the high workload scenarios. The prolonged occurrence of off-nominal events could contribute to a slightly heightened level of complexity. Additionally, the deviations in communication patterns can differ across Air Traffic Control Officers (ATCos), possibly due to variations in individual experience and seniority. Therefore, it can be concluded that both airspace density and off-nominal events contribute to an increase in communication complexity. As depicted in Figure \ref{fig: hitl-eval}(b), it is evident that the correlation, predictability, and stability of the system all exhibit a decrease from the baseline scenarios to the high workload scenarios. These findings provide valuable insights into understanding the behavior of different scenarios and guide our further studies on workload prediction.

\section{Proposed ATC Workload Prediction Framework \label{sec: methodologies}}
\begin{figure}[H]
    \centering
    \includegraphics[width=\textwidth]{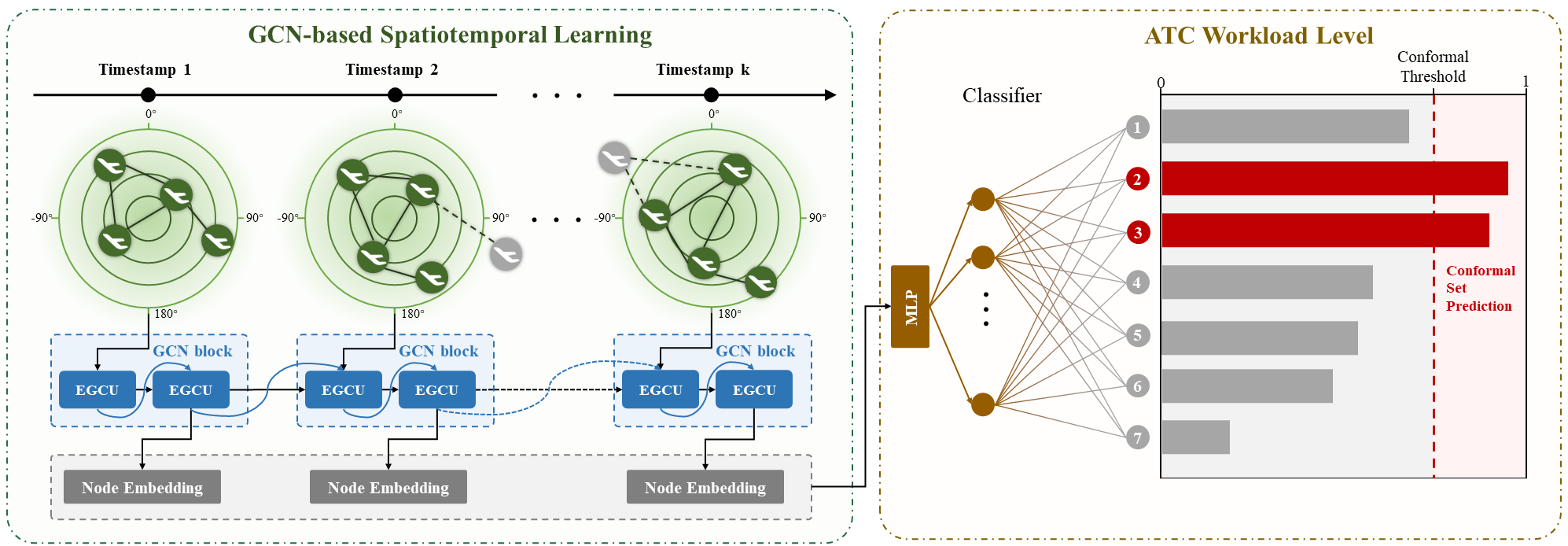}
    \caption{Schematic illustration of conformalized EvolveGCN set prediction framework. We formulate the ATC workload level prediction as a time-series graph classification task, where each graph node represents \textit{each aircraft under the ATC's control}. The number of nodes and weight (distance) of each edge can change across different timestamps. On the classifier side, we propose the conformal prediction set for improved ground truth coverage. Conformalization acts as a \textit{post-hoc} procedure to post-process the prediction labels, where the softmax probability threshold is inferred on the calibration set.}
    \label{fig: gcn-flowchart}
\end{figure}

In this section, we describe the proposed workload machine learning prediction framework in \Cref{fig: gcn-flowchart}. We first demonstrate the formulation and basic concepts of the graph learning problems in \Cref{subsec: problem_formulation}. The dynamical graph convolution learning algorithm and conformal prediction setup are explained in \Cref{subsec: egcn} and \Cref{subsec: cp}, respectively.

\subsection{Problem Formulation} \label{subsec: problem_formulation} 

As mentioned, the ATCo workload prediction task is viewed as a time-series dynamical graph classification task. In this paper, we use subscript $t \in \{1, ..., T\}$ to demonstrate the timestamp and superscript $l \in \{1, ..., L\}$ to denote the layer index. Graph neural networks (GNNs) are introduced to model the spatiotemporal layout of the airspace from the structured graph data with explicit message passing \citep{li2019grip, mohamed2020social}. We denote a graph at timestamp $t$ with vertices and edges, represented as $G_t = (V_t, E_t)$, where the number of nodes at timestamp $t$ is $N_t^{nodes} = |V_t|$ and the number of edges at timestamp $t$ is $N_t^{edges} = |E_t|$. The adjacency matrix $A_t \in \mathbb{R}^{N_t^{nodes}\times N_t^{nodes}}$. The constructed graph structure can be either directed or undirected depending on whether the edges are directed from one node to another. The dynamic graph is mentioned when the graph topology varies with time. Especially, the graphs in our work are undirected dynamical graphs. The constructed graph inputs are $A_t \in \mathbb{R}^{N_t^{nodes}\times N_t^{nodes}}$ and $X_t \in \mathbb{R}^{N_t^{nodes}\times N_t^{features}}$, where $A_t$ is the adjacency matrix at each timestamp t and $X_t$ is the node feature matrix.

\begin{figure}[H]
    \centering
    \includegraphics[width=0.75\textwidth]{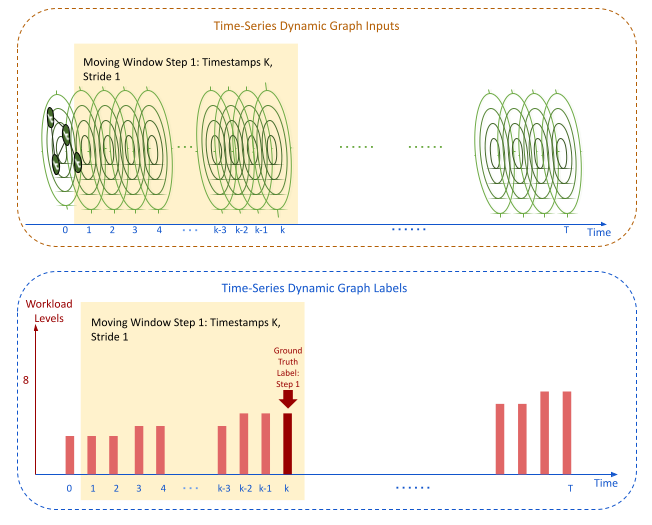}
    \caption{Schematic illustration of the moving window approach. At each step, the moving window moves 1 timestamp (5s) along the temporal dimension (stride 1). Each series of graphs contains a graph of $\kappa$ timestamps. The workload ground truth label for the graph series input is the workload level (1-7) reported at the last timestamp, collected from the human-in-the-loop experiment. This setup allows the model to capture long-term spatial relationships and result in a prediction at every timestamp since $\kappa$. For abbreviation, the graph input is represented by a radar plot.}
    \label{fig: window}
\end{figure}

Graph structure represents the spatial layout of the airspace. However, the dynamical graph constructed $A_t$ is based on the geo-distance between two aircraft pairs, resulting in geospatial graphs on a two-dimensional space. This type of graph construction is widely adopted with acceptable complexity. Consequently, in this work, we adopt a scaling formula described in minimum-spanning-tree-based workload prediction task \citep{sridhar1998airspace} to scale the horizontal and vertical distance between two aircraft pairs into one distance metric. Specifically, the graph is built by calculating the distance $\tilde{d_{ij}}$ between two aircraft pairs $(i, j)$ at the same timestamp $t$. We use $d_{t, ij}$ to represent the horizontal distance and $h_{t, ij}$ to present the vertical separation distance between aircraft pairs $(i, j)$. The scaling function is described in \Cref{{eq: scale_distance}}.

\begin{equation} \label{eq: scale_distance}
    \tilde{d_{t, ij}} = \sqrt{d^2_{t, ij} + s^2 h^2_{t, ij}}
\end{equation}

\noindent where $s$ is the spatial scaling factor which equalizes the separation on the horizontal and vertical dimension, as in \Cref{eq: scale_factor}. 

\begin{equation} \label{eq: scale_factor}
     s = \left\{ \begin{array}{rcl}
0.005, & \mbox{for} & \mathsf{alt}_i \leq 29,000 \hspace{0.25em} \mathsf{and} \hspace{0.25em} \mathsf{alt}_j \leq 29,000  \\ 
0.0025, & \mbox{for} & \mathsf{alt}_i > 29,000 \hspace{0.25em} \mathsf{or} \hspace{0.25em} \mathsf{alt}_j > 29,000 \\
\end{array}\right. i,j = 1, ..., N
\end{equation}

In this workload prediction task, we first obtain the constructed input $X_t$ and $A_t$ at each timestamp $t$. In such a way, we obtain the series of graphs $G_t, t \in \{1, ..., T\}$. Then, we fill the workload ratings into another time series based on the self-evaluated workload rating during the HITL simulations, representing the prediction labels. Last, we use a moving window approach to build the correct input-output matching for supervised machine learning. The schematic illustration of the moving window process is shown in \Cref{fig: window}. We define a window of size $\kappa$ and move the window along the time axis of the series of inputs, with a stride of 1. The time-series graph of size $\kappa$ is denoted as $\{G_t\}_{\kappa}$. Similarly, we move the window function along the prediction labels and obtain the workload ratings by claiming the last reported workload value within the current window. Then, if $Y_t$ denotes the ground truth workload label at timestamp $t$, we mathematically formulate the problem into,

\begin{equation}
    Y_t = \mathsf{EvolveGCN}(\{G_t\}_{\kappa})
\end{equation}

\subsection{Evolving Graph Convolution Network \label{subsec: egcn}}

Spatial graph convolutional networks (GCN) \citep{kipf2016semi} convolve the input $A_t$ and $X_t$ using the derived compact form,

\begin{equation}
    H_t^{l+1} = \sigma (\hat{A_t} H_t^l W_t^l), \quad with \quad H_t^0 = X_t
\end{equation}

\noindent where $\sigma$ is the activation function (i.e., $\mathsf{ReLU}$). $\hat{A_t}$ is a normalized version of $A_t$, to account of numerical instability. Specially, $\hat{A_t}$ is defined as, $\hat{A_t} = \tilde{D_t}^{-\frac{1}{2}} \tilde{A_t} \tilde{D_t}^{-\frac{1}{2}}$, $\tilde{A_t} = A_t + I_t$, $\tilde{D}_{t, ii} = \sum_j \tilde{A}_{t, ij}$. It's clear that $H_t$ has the same dimension as $X_t$ as $H_t \in \mathbb{R}^{N_t^{nodes}\times N_t^{features}}$. $W_t \in \mathbb{R}^{N_{features}\times N_{featrues}}$ is the kernel parameters. For multiple graph convolutional layer setup, $H_t^{l+1}$ stands for the updated graph embedding of convolutional layer $l+1$ at timestamp $t$. Specifically, in classification problems, the activation function $\sigma$ at the output layer $L$ is the $\mathsf{softmax}$ function. 

Evolving Graph Convolution Network (EvolveGCN) improves GCN by introducing recurrence layers to capture the dynamism underlying a time-series graph. Two types of EvolveGCN are presented \citep{pareja2020evolvegcn}, depending on the recurrent updating architecture.

The first variant treats the GCN kernel parameter $W_t^l$ as the hidden state of recurrent learning function and updates $W_t^l$ with a gated recurrent unit (GRU), while the node embeddings of node features are still contained within the GCN hidden state tensor $H_t^l$. EvolveGCN-H is used to denote this variant \Cref{eq: egcn-h}. This requires a special design of GRU computation flows as described in \citep{pareja2020evolvegcn}.

\begin{align}\label{eq: egcn-h}
    W_t^l &= GRU(H_t^l, W_{t-1}^l)\\
    H_t^{l+1} &= \sigma (\hat{A_t} H_t^l W_t^l)
\end{align}

Another variant of EvolveGCN is the -O version \Cref{eq: egcn-o}, where the kernel parameter $W_t^l$ is treated as input of recurrent learning without considering the temporal correlations between node embeddings. The implementation of EvolveGCN-O is straightforward by extending dimensions.

\begin{align}\label{eq: egcn-o}
    W_t^l &= LSTM(W_{t-1}^l)\\
    H_t^{l+1} &= \sigma (\hat{A_t} H_t^l W_t^l)
\end{align}

Either an EvolveGCN-H or EvolveGCN-O is denoted as an Evolving Graph Convolution Unit (EGCU), as shown in \Cref{fig: gcn-flowchart}. In both ways, the EGCU first updates the GCN weights and then propagates the hidden states through the layers. Several layers of EGCU form a GCN block in \Cref{fig: gcn-flowchart}. For a graph learning problem with large feature space, EvolveGCN-H is more effective since the feature embedding recurrence is also considered. Otherwise, EvolveGCN-O is more focused on learning the graph topology structure changes.

\subsection{Conformal Prediction \label{subsec: cp}}
In this section, we provide a brief overview of conformal prediction (CP). For the classification task mentioned above, we have EvolveGCN acted as the classifier $\mathcal{C}$, which outputs an estimated probability for each class, i.e., $p \in [0,1)^\varkappa$ for $\varkappa$ classes. We reserve a small amount of data called \textit{calibration set} to calculate the probability score threshold $\hat{q}$ such that the following condition holds on the test set,

\begin{equation}
    1-\alpha \leq \mathbb{P}(Y_{test} \in \mathcal{C}(X_{test})) \leq 1-\alpha+\frac{1}{n+1}
\end{equation}

 \noindent where the test dataset is the unused test set to evaluate model performance. $\alpha \in (0, 1]$ is the pre-defined tolerated error rate. This is to guarantee that the model is $1-\alpha$ confident that the model prediction set contains the correct ground truth label. This equation is also known as the marginal conformal coverage guarantee, which has been proved in the literature \citep{vovk1999machine,angelopoulos2021gentle}. Notably, the \textit{calibration} is the key step to find $\hat{q}$. Suppose we define the concept of the conformal score by one minus the softmax probability of the true class, $\hat{q}$ is defined to be the $\frac{\lceil(n+1)(1-\alpha)\rceil}{n}$ quantile of the conformal scores. $\lceil \rceil$ is the ceiling function to correct the quantile. Then, the prediction of a new test sample will be all classes with a softmax score higher than $\hat{q}$. The prediction set will be larger if the model is uncertain about the prediction labels or if the input is out-of-distribution. Intrinsically, the size of the prediction set is the indicator of model uncertainty. 

 Conformal prediction (CP) has been studied from various angles by researchers. However, the classical CP method is susceptible to coverage issues due to its tendency to produce the smallest average size of prediction sets \citep{sadinle2019least}. Specifically, CP tends to overcover hard data samples while undercover simple ones. To address this issue, researchers have proposed an approach called adaptive conformal prediction \citep{angelopoulos2020uncertainty,romano2020classification}. The underlying principle of this method is to compute the conformal threshold $\hat{q}$ based on the cumulative softmax score across $\varkappa$ classes.

It's also noteworthy to discuss conformal evaluation methods, which are adopted in evaluating our model. To determine the model's performance, a straightforward method is to examine the histogram of prediction set sizes visually. Essentially, a larger size of the prediction set implies that the model is facing certain data quality problems, while the variation in the set size can provide insights into the model's ability to differentiate between easy and difficult input samples.

 \begin{equation} \label{eq: conditional_coverage}
     \mathbb{P}[Y_{test} \in \mathcal{C}(X_{test}) | X_{test}] \geq 1-\alpha
 \end{equation}
 
Conditional coverage is a feasible approach to evaluate the adaptivity of conformal prediction. For instance, in a classification setting, we seek to find the prediction sets with exactly $1-\alpha$ coverage for any input data sample, as in \Cref{eq: conditional_coverage}. The conditional coverage concept is a stronger metric than the marginal coverage mentioned above. Some literature mentioned that conditional coverage is impossible to achieve in most general cases \citep{angelopoulos2022conformal}. \textbf{Size-stratified coverage (SSC) metric} is a general metric to evaluate how close the model is able to achieve \Cref{eq: conditional_coverage}. SSC metric is a way to evaluate the performance of conformal prediction models. It is based on the idea that prediction sets of different sizes may have different properties and should be evaluated separately. The key is to group test samples into different size strata based on the size of their prediction sets and compute the averaged empirical coverage on each size strata. SSC metric can be useful to diagnose specific issues, such as overcoverage or undercoverage, that may be related to the size of the prediction sets. In addition, another conformal prediction evaluation method has been proposed recently \citep{olsson2022estimating}, where a calibration plot of the prediction error versus the specified significance level ($\alpha$) is used. Remarkably, In \Cref{sec: experiments}, we evaluate our model with these metrics.

CP provides a rigorous way to measure the uncertainty associated with the predictions made by a machine learning model and to express this uncertainty in the form of prediction intervals or regions that can be used to guide decision-making. This can be particularly useful in critical engineering applications where accurate prediction intervals or regions are essential \citep{balasubramanian2014conformal}. CP has been widely adopted in drug discovery \citep{alvarsson2021predicting}, medical diagnosis \citep{lu2022fair}, and robotics \citep{luo2022sample}. CP is a unified \textit{post-hoc} softmax score calibration process to generate prediction sets for any classification model \citep{papadopoulos2002inductive, vovk2005algorithmic, lei2014distribution}. In this work, we propose to use CP for aviation decision support. Specifically, the prediction set comes from CP gives uncertain prediction label suggestions, i.e., workload rating of 3, 5, 7. While the isolated classification label in workload prediction is not reasonable, we propose to fill the intermediate workload ratings based on the minimum predicted rating and the maximum predicted ratings.

\section{Experiments \label{sec: experiments}}
In previous sections, we have introduced the HITL data collection process, the problem definition, the machine learning model system design, and conformal prediction for better ground truth label coverage \Cref{sec: methodologies}. In this section, we present a comprehensive of experiments to test and evaluate the proposed model. We first discuss several evaluation metrics used for this classification task. Then we report the classification accuracy from the machine learning model with a few implementation details. Lastly, conformal prediction set results are reported with several conformal coverage evaluation methods mentioned in \Cref{subsec: cp}. The validation set is used to tune hyper-parameters, and the testing dataset results are evaluated based on the best validation epoch.

\subsection{Evaluation Metrics \label{subsec: metrics}}
The F-score, or F-measure, is a binary classification metric used in statistical analysis to assess the accuracy of test samples. Specifically, the F1-score is defined as the symmetrical harmonic mean of precision and recall \citep{taha2015metrics}. F1-score can also be used for multi-class classification by taking either the micro-averaging (MicroF1) or the macro-averaging (MacroF1).  


\begin{align} \label{eq: microf1score}
    \mathsf{Precision} &= \frac{\mathsf{TP}}{\mathsf{TP}+\mathsf{FP}} \\
    \mathsf{Recall} &= \frac{\mathsf{TP}}{\mathsf{TP}+\mathsf{FN}} \\
    \mathsf{MicroF1} &= 2*\frac{\mathsf{Precision}*\mathsf{Recall}}{\mathsf{Precision}+\mathsf{Recall}} \\
\end{align}


\Cref{eq: microf1score} gives the mathematical formulation of MicroF1, where each sample is considered independently without considering which class this sample belongs to. MicroF1 treats each data input equally but is biased on class frequency. MicroF1 is useful when the classification task is unbalanced, meaning that some classes have many more instances than others. In this case, the MicroF1 gives equal weight to each instance, regardless of its class. 
 
\begin{equation} \label{eq: macrof1score}
    \mathsf{MacroF1} = \frac{\mathsf{F1_1} + \mathsf{F1_2} \cdots , + \mathsf{F1_n}}{\mathsf{n}}
\end{equation}

On the contrary, MacroF1 average the F1 score across all classes as in \Cref{eq: macrof1score}, where $n$ is the number of classes, and $\mathsf{F1_1} + \mathsf{F1_2} \cdots , + \mathsf{F1_n}$ are the F1 scores for each class. MacroF1 treats each class equally regardless of the size of samples within each class thus, it's biased on the number of samples. MacroF1 is useful when the classification task is balanced, meaning that each class has approximately the same number of instances. In this case, MacroF1 gives equal weight to each class, regardless of its frequency.

In this work, we are encountering a highly-imbalanced classification problem \Cref{fig: collecteddata}. Thus, MicroF1 is a better indicator than MacroF1. It is noteworthy to mention that $n$ in \label{eq: macrof1score} should be adjusted to exclude the class labels that are not presented in either ground truth or predictions. We report both metrics in our studies.

\subsection{Implementation Details \label{subsec: details-workload}}
Although the commonly adopted node-level or link-level classification objective is prevalent, the proposed workload prediction framework is instead a graph-level classification task \citep{pareja2020evolvegcn}. Thus, on the output layer, we take the aggregated class probability score across each node to get a unified score of the entire graph.  We adopted the grid-search strategy to search for key parameters and fine-tuned the neural network model with the data collected from three different scenarios. The key parameters used are the number of EvolveGCN layers (EGCU layers), and the dimensions of these layers. Dropout is used in the classifier to address overfitting. \Cref{table: egcu-parameters} lists the fine-tuned key model parameters under three simulation scenarios. Other parameters, such as the dimension of classifiers, are kept the same as the original implementation \citep{pareja2020evolvegcn}. Parameter tuning on the classifiers might be useful, but it's beyond the scope of this study. 

\begin{table}[htb]
\caption{List of fine-tuned model parameters used in EvolveGCN training under three different simulation scenarios.}
\label{table: egcu-parameters}
\centering
\begin{tabular}{c|c|c|c}
\hline
                      & Baseline & \begin{tabular}[c]{@{}c@{}}High \\ Workload \\ Nominal\end{tabular} & \begin{tabular}[c]{@{}c@{}}High \\ Workload \\ Off-Nominal\end{tabular} \\ \hline
Number of EGCU Layers & 2        & 2                                                                   & 4                                                                       \\ \hline
EGCU Layer Dimensions & 64       & 128                                                                 & 64                                                                      \\ \hline
Dropout Ratio         & 0.25     & 0.5                                                                 & 0.25                                                                    \\ \hline
Learning Rate         & 0.001    & 0.0015                                                              & 0.0005                                                                  \\ \hline
\end{tabular}
\end{table}

Moreover, as discussed in \Cref{sec: human}, the first reported workload rating starts at 3 minutes of the 25 minutes duration. This corresponds to the $36th$ timestamp with $5s$ interval in the collected flight traffic data. Consequently, the moving window size $\kappa$ in our experiment is 36, with a stride of 1. We separate the data into train, validation, and test sets with a ratio of $[0.4, 0.3, 0.3]$. The validation set is used for deep learning model hyper-parameter tuning. Moreover, the validation set is also used as the calibration set to find the CP threshold $\hat{q}$.

\subsection{Experiment Results \label{subsec: sota-workload}}

\begin{table}[htb]
\caption{Workload Level Prediction: Comparison between different workload prediction methods.}
\label{table: results}
\resizebox{\textwidth}{!}{
\begin{tabular}{c|cc|cc|cc}
\hline
\multirow{2}{*}{\textbf{\begin{tabular}[c]{@{}c@{}}ATC Workload \\ Level Prediction\end{tabular}}} & \multicolumn{2}{c|}{\textbf{Baseline}}               & \multicolumn{2}{c|}{\textbf{High Workload Nominal}}  & \multicolumn{2}{c}{\textbf{High Workload Off-Nominal}} \\ \cline{2-7} 
                                                                                                   & \multicolumn{1}{c|}{MicroF1}        & MacroF1        & \multicolumn{1}{c|}{MicroF1}        & MacroF1        & \multicolumn{1}{c|}{MicroF1}         & MacroF1         \\ \hline
\textit{\begin{tabular}[c]{@{}c@{}}Simple LR\\ w/ Density\end{tabular}}                            & \multicolumn{1}{c|}{0.306}          & 0.218          & \multicolumn{1}{c|}{0.307}          & 0.198          & \multicolumn{1}{c|}{0.323}           & 0.195           \\ \hline
\textit{\begin{tabular}[c]{@{}c@{}}Simple LR\\ w/ Graph Feature\end{tabular}}                      & \multicolumn{1}{c|}{0.331}          & 0.236          & \multicolumn{1}{c|}{0.383}          & 0.263          & \multicolumn{1}{c|}{0.350}           & 0.255           \\ \hline
\textit{\begin{tabular}[c]{@{}c@{}}2-layer\\ MLP\end{tabular}}                                     & \multicolumn{1}{c|}{0.364}          & \textbf{0.283} & \multicolumn{1}{c|}{0.455}          & 0.386          & \multicolumn{1}{c|}{0.459}           & 0.367           \\ \hline
\textit{GCN}                                                                                       & \multicolumn{1}{c|}{0.404}          & 0.218          & \multicolumn{1}{c|}{0.580}          & 0.401          & \multicolumn{1}{c|}{0.526}           & 0.352           \\ \hline
\textit{EvolveGCN-O}                                                                               & \multicolumn{1}{c|}{\textbf{0.545}} & 0.277          & \multicolumn{1}{c|}{\textbf{0.740}} & \textbf{0.632} & \multicolumn{1}{c|}{\textbf{0.695}}  & \textbf{0.472}  \\ \hline
\textit{EvolveGCN-H}                                                                               & \multicolumn{1}{c|}{0.413}          & 0.221          & \multicolumn{1}{c|}{0.593}          & 0.474          & \multicolumn{1}{c|}{0.581}           & 0.414           \\ \hline
\end{tabular}

}
\end{table}

In this work, we compare our model with both classical methods (i.e., linear regressions (LRs)) and simple data-driven learning methods (i.e., fully-connected neural networks/multilayer perceptrons (MLPs)). In \cite{hah2006effect}, the authors also conducted high-fidelity human-in-the-loop simulations to study the impact of traffic density features on controller workload. They found that the workload rating of the enroute center controller is proportional to the number of aircraft with a slope of $0.306$ and bias of $-3.373$. They also identified the primary sources of workload for controllers, including airspace and traffic management, communication, and coordination tasks with workload management suggestions. In \cite{sridhar1998airspace, chatterji2001measures}, the authors create graph-structured airspace data structure -- minimum-spanning trees but propose several handcraft features based on the histogram of node features. Then a two-layer fully-connected neural network is used for prediction based on handcrafted features and shows remarkable performance. However, the workload ratings are directly generated from the traffic density, where thresholds of 7 and 17 separate workload ratings into low, medium, and high scenarios. Likewise, inspired by this work, our proposed method adopts a graph structure to represent the spatiotemporal layout. We utilize the recent advancement in graph learning and learn from the graph structure without handcrafted features. 

In \Cref{table: results}, comparing the first two rows, we first show that including additional graph node features can achieve higher prediction accuracy, even for simple LRs. Despite the traffic density features, additional graph node features are traffic conflict features (i.e., horizontal/vertical minimum separation to nearby aircraft). For the MLP with handcraft features, we generate second-order statistics of the sum and difference histograms introduced in \citep{sridhar1998airspace}. As a reference to EvolveGCN, we also conduct an experiment on vanilla GCN. This can be easily achieved by removing the LSTM layer in \Cref{eq: egcn-o}. We show that EvolveGCN can achieve significantly higher MicroF1 and MacroF1 than LP, MLP, and GCN. Moreover, the -O variant EvolveGCN outperforms the -H variant. One of the major reasons is the selection of top K indices reduces the hidden dimension of EGCU, due to the low node feature dimensions and a small number of nodes in graphs (i.e., only one aircraft showing up at the first timestamp) at certain timestamps.

\subsection{Conformal Prediction Results}
\begin{figure}[H]
    \centering
    \begin{subfigure}[t]{0.75\textwidth}
        \centering
        \includegraphics[width=\textwidth]{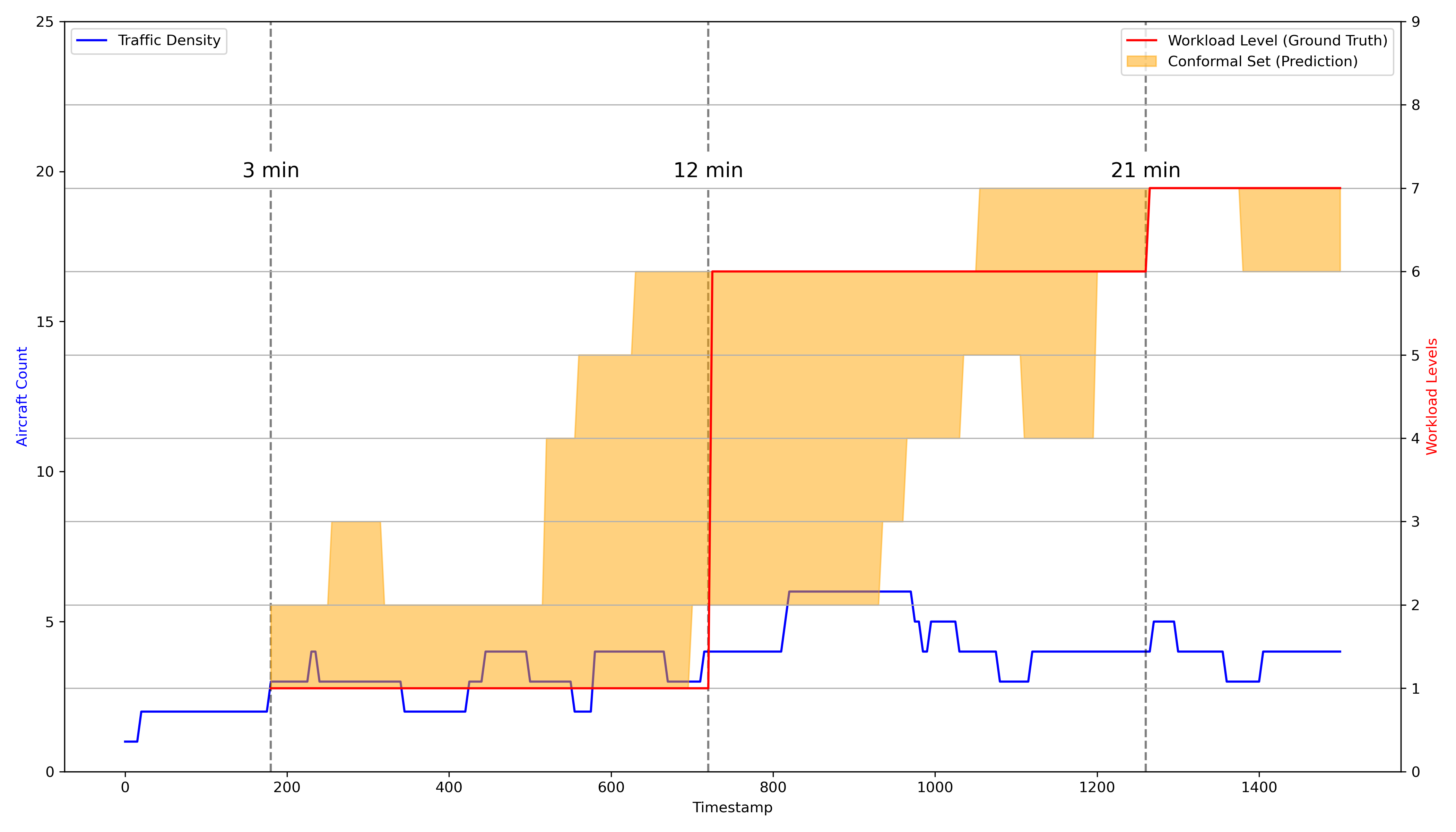}
        \caption{Baseline Condition}
        \label{fig: baselinecondition} 
    \end{subfigure}
    ~
    \begin{subfigure}[t]{0.75\textwidth}
        \centering
        \includegraphics[width=\textwidth]{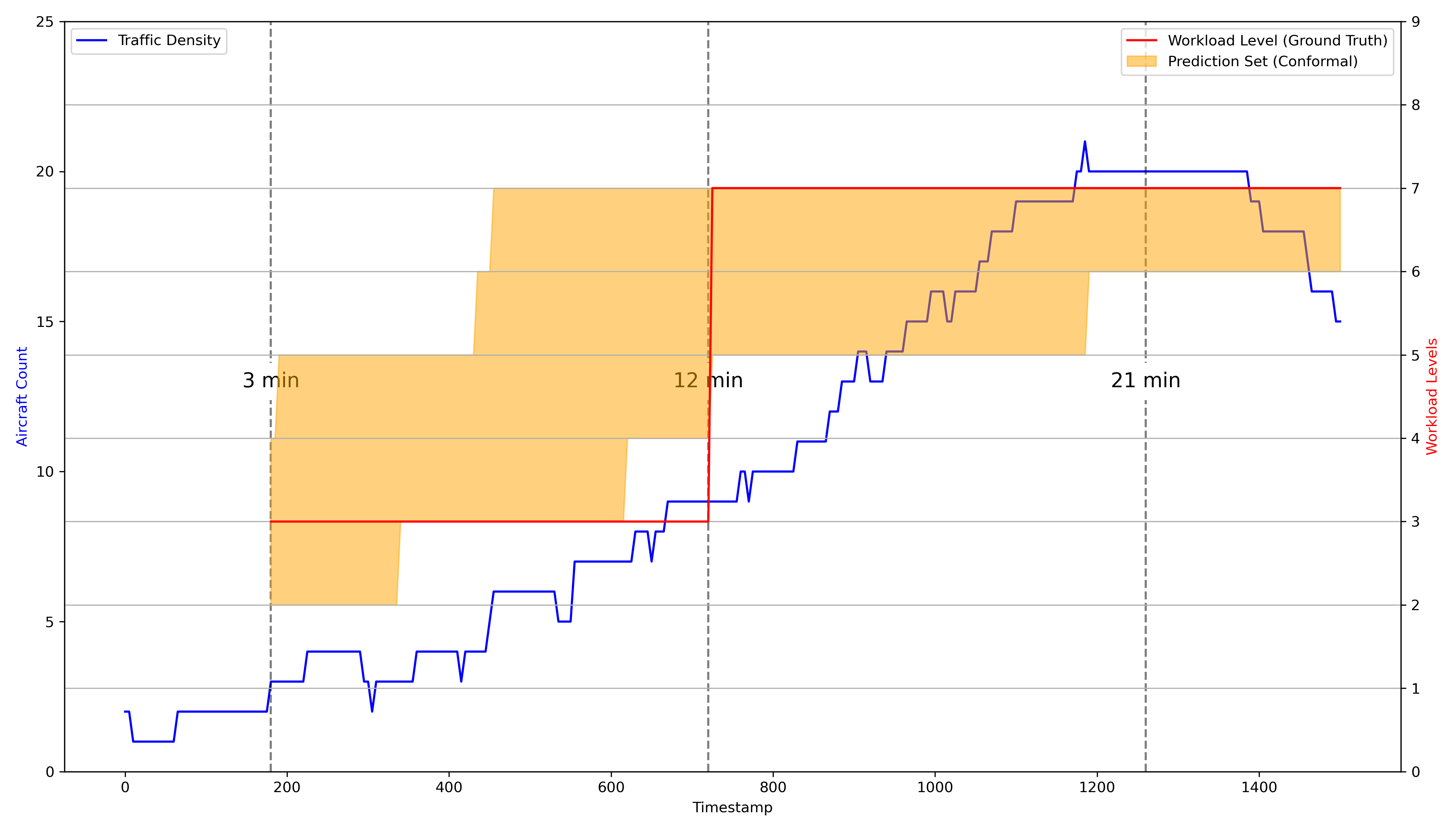}
        \caption{High Workload Nominal Condition}
        \label{fig: nominalcondition} 
    \end{subfigure}
    ~
    \begin{subfigure}[t]{0.75\textwidth}
        \centering
        \includegraphics[width=\textwidth]{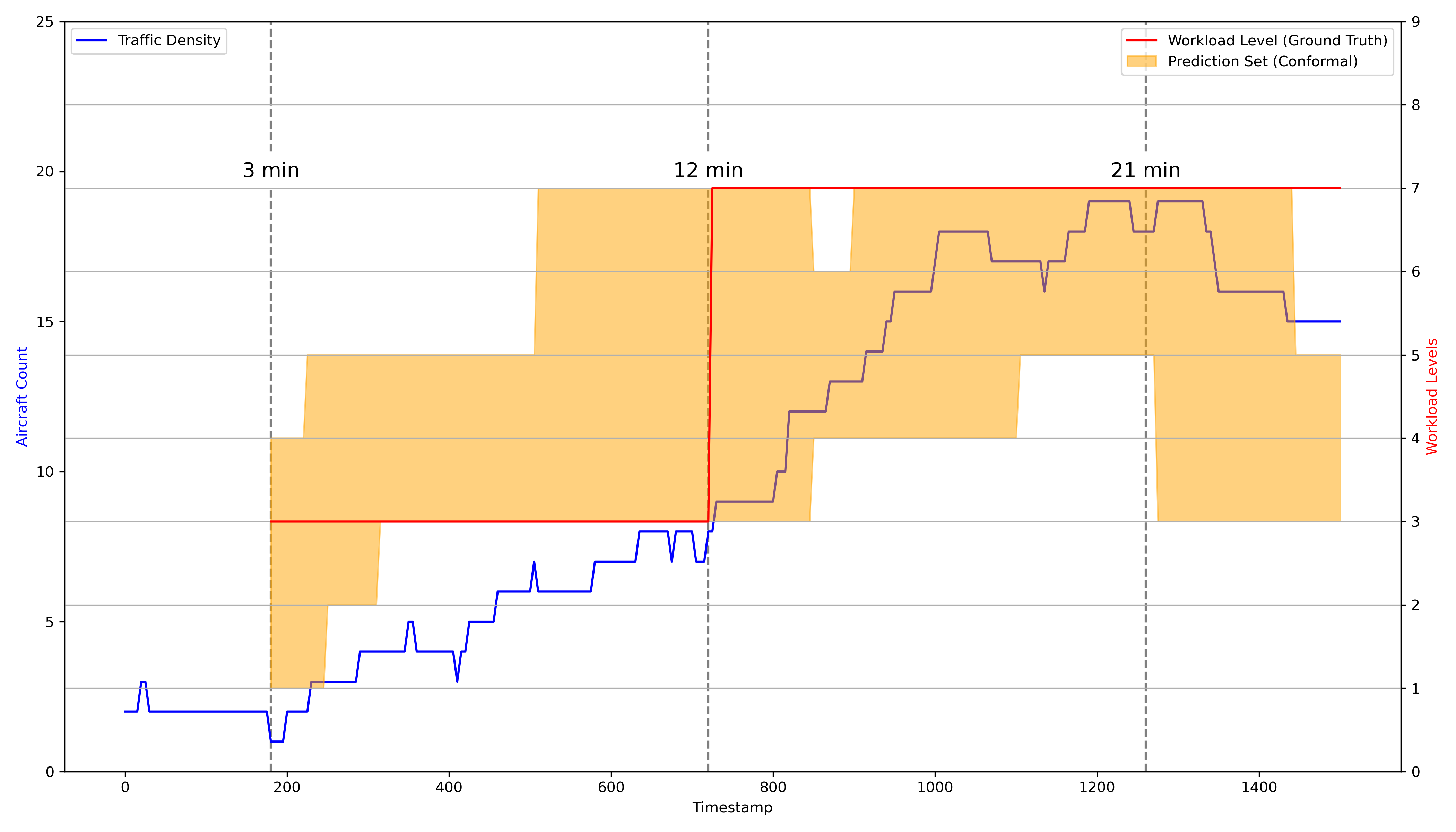}
        \caption{High Workload Off-Nominal Condition}
        \label{fig: offcondition} 
    \end{subfigure}
    \caption{Visualization of conformal predictions on the test sample. For the workload level prediction task, we set our prediction as the range between the lowest predicted workload level and the highest predicted workload level.}
    \label{fig: pred-vis}
\end{figure}

As mentioned, we use conformal prediction to improve prediction accuracy further. In \Cref{fig: pred-vis}, the prediction on one test participant is shown. The conformal prediction set coverage is the shaded region. The conformal prediction is generated with a tolerated error rate of $\alpha = 5\%$. The blue solid lines show the real-time aircraft density (left axis) in the simulation, and the red lines are the interpolated workload ratings. At 3 min, 12 min, and 21 min, the participants are required to submit their workload rating to the computer. The ground truth workload ratings are colored in red (right axis). The conformal prediction set covers most of the ground truth but is undercover at several spots. As discussed in \Cref{subsec: cp}, we further evaluate the conformal prediction coverage.

\subsection{Conformal Coverage Evaluation}
\begin{figure}[H]
    \centering
    \begin{subfigure}[t]{0.45\textwidth}
        \centering
        \includegraphics[width=\textwidth]{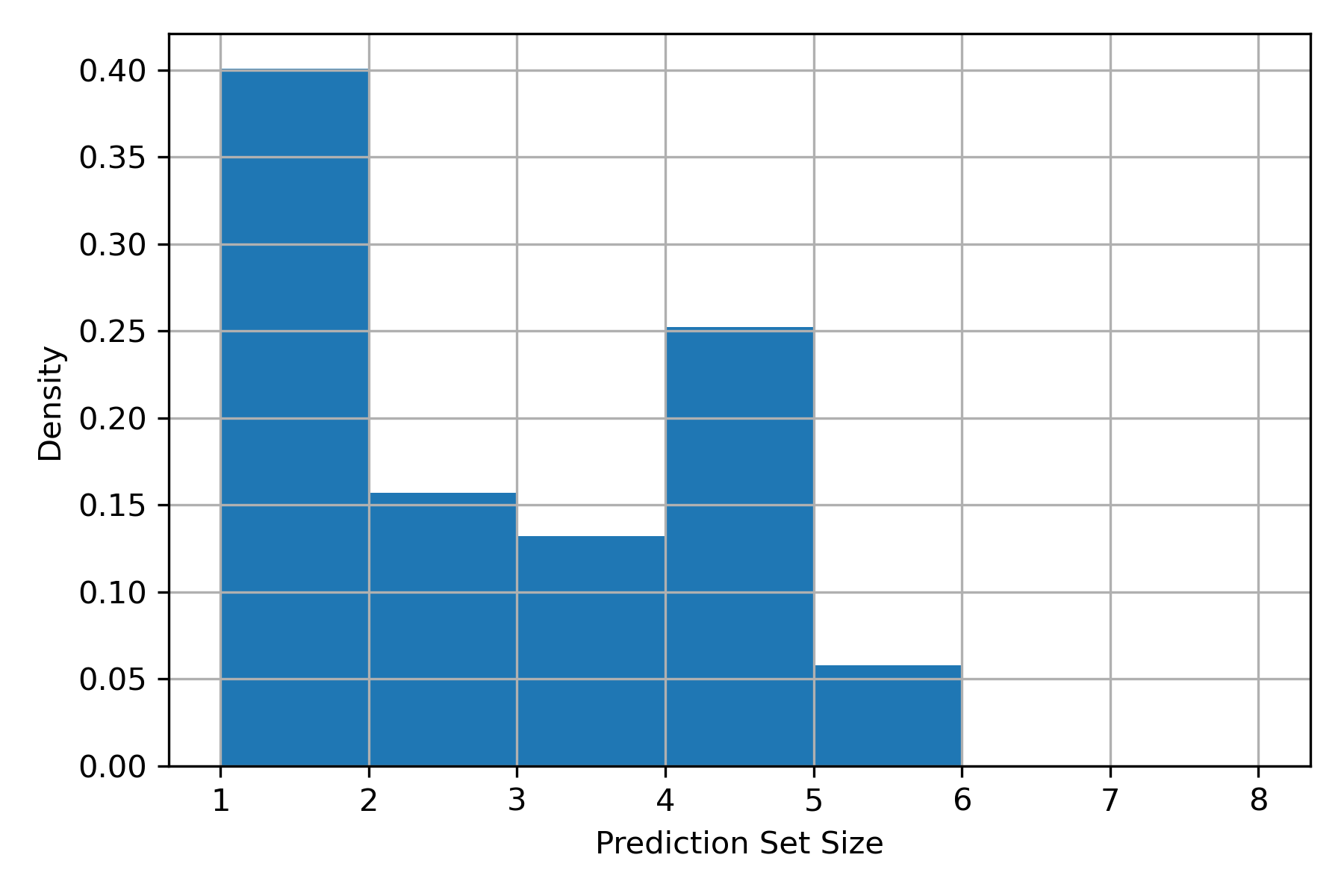}
        \caption{Baseline Condition}
        \label{fig: eval-baselinecondition} 
    \end{subfigure}
    ~
    \begin{subfigure}[t]{0.45\textwidth}
        \centering
        \includegraphics[width=\textwidth]{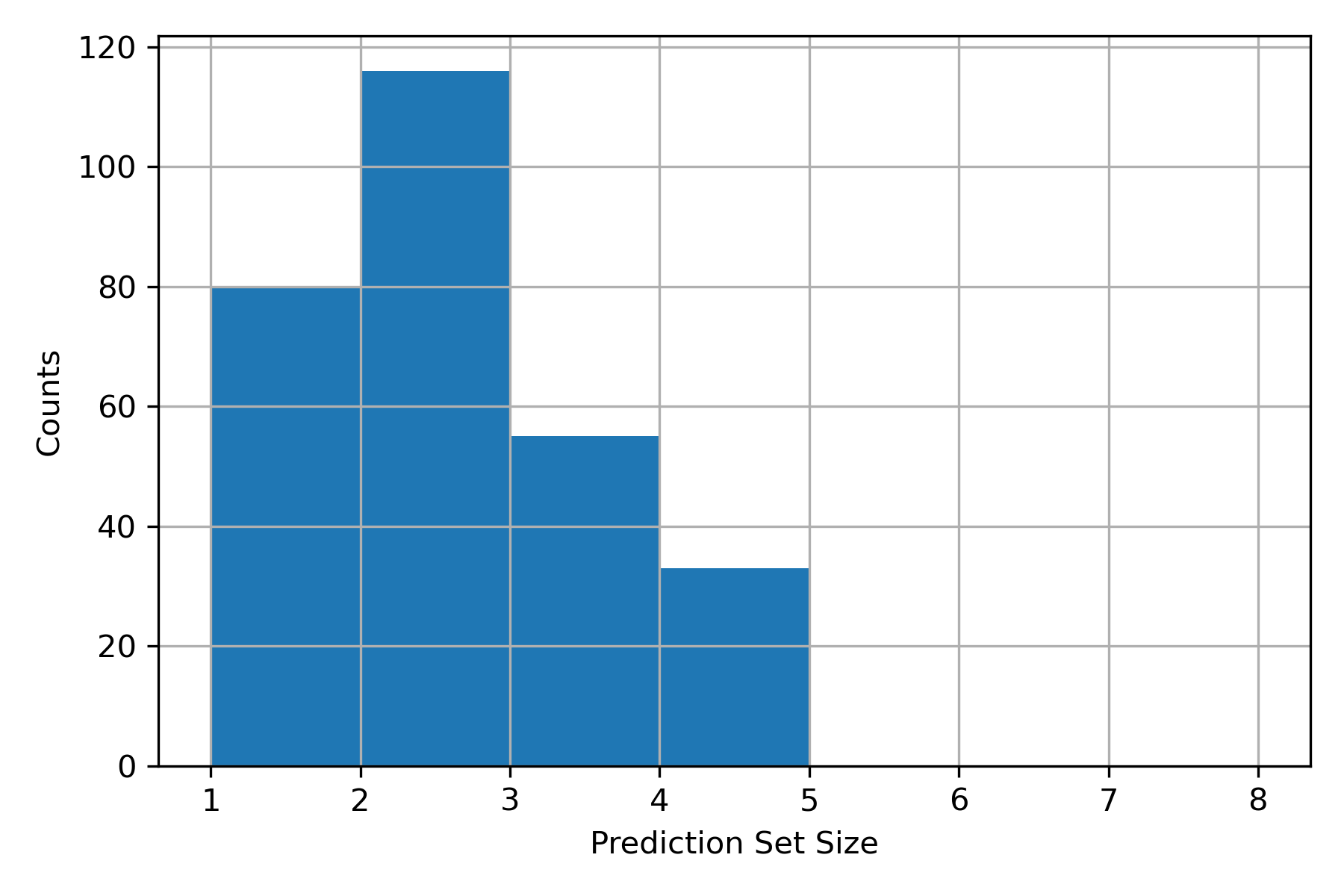}
        \caption{High Workload Nominal Condition}
        \label{fig: eval-nominalcondition} 
    \end{subfigure}
    ~
    \begin{subfigure}[t]{0.45\textwidth}
        \centering
        \includegraphics[width=\textwidth]{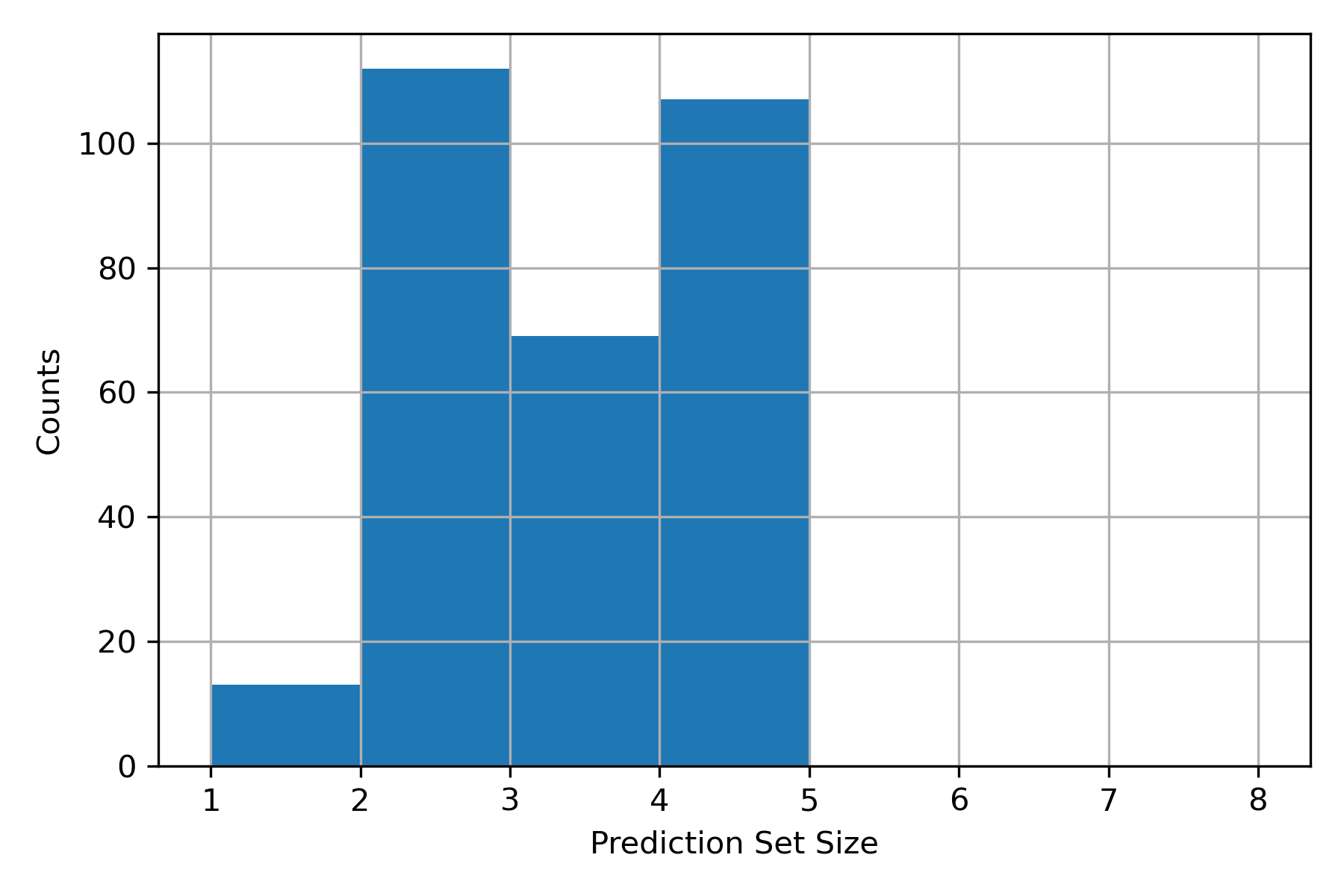}
        \caption{High Workload Off-Nominal Condition}
        \label{fig: eval-offcondition} 
    \end{subfigure}
    \caption{Histogram of set sizes on test set predictions. The spread of the histogram shows the difficulty of making a correct prediction.}
    \label{fig: setsize-vis}
\end{figure}

We adopt different conformal coverage evaluation metrics to examine the performance of our conformal set. Firstly, we plot the histogram of set sizes. A high average set size suggests that the conformal prediction procedure is imprecise, which could indicate issues with the score or underlying model. Secondly, the range of set sizes indicates whether the prediction sets adapt properly to the complexity of examples. A wider range is typically preferred because it implies that the procedure accurately differentiates between simple and challenging inputs. We show the histograms in \Cref{fig: setsize-vis}. The size of conformal prediction is typically around 5. The spread for baseline and high workload nominal conditions looks reasonable. The model is able to distinguish hard and easy samples. However, the spread for high workload off-nominal conditions indicates potential scoring issues or simply difficult data \citep{angelopoulos2021gentle}. 

\begin{figure}[H]
    \centering
    \includegraphics[width=\textwidth]{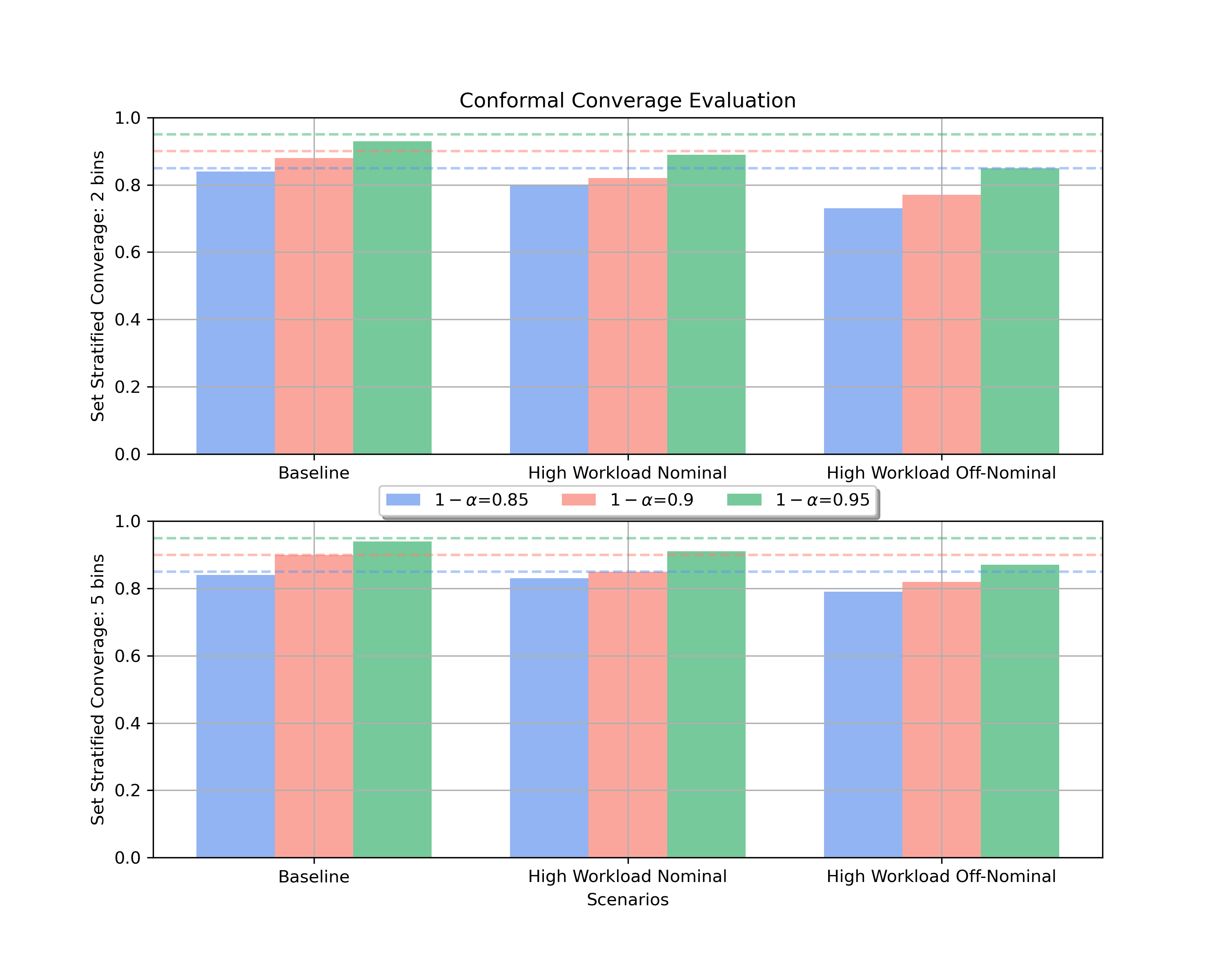}
    \caption{Conformal coverage evaluation with various desired $\alpha$ values under three workload simulation conditions. We use the Size-Stratified coverage (SSC) metric better to represent the adaptive coverage of the conformal set coverage.}
    \label{fig: setcoverage-vis}
\end{figure}

Following the discussion in \Cref{subsec: cp}, we investigate the size-stratified coverage (SSC) to evaluate the condition coverage and plot the figures in \Cref{fig: setcoverage-vis}. The dash lines are desired coverage values. In this figure, we consider three desired error rates, $\alpha=0.15, 0.1, 0.05$, for three scenarios. We use two possible $\mathcal{C}(x)$ cardinalities of two bins and five bins. In other words, we divide the predicted sets into different size categories (e.g., sets of size 2, sets of size 5.) and calculate the percentage of times that the true value falls within each category. We discover that the prediction coverage of baseline conditions shows a good sign, but the two high workload scenarios tend to under-coverage. Again, the reasons still come from the unsatisfactory of the collected data, which leads to lower reported F1 values. 

\begin{figure}[H]
    \centering
    \begin{subfigure}[t]{0.45\textwidth}
        \centering
        \includegraphics[width=\textwidth]{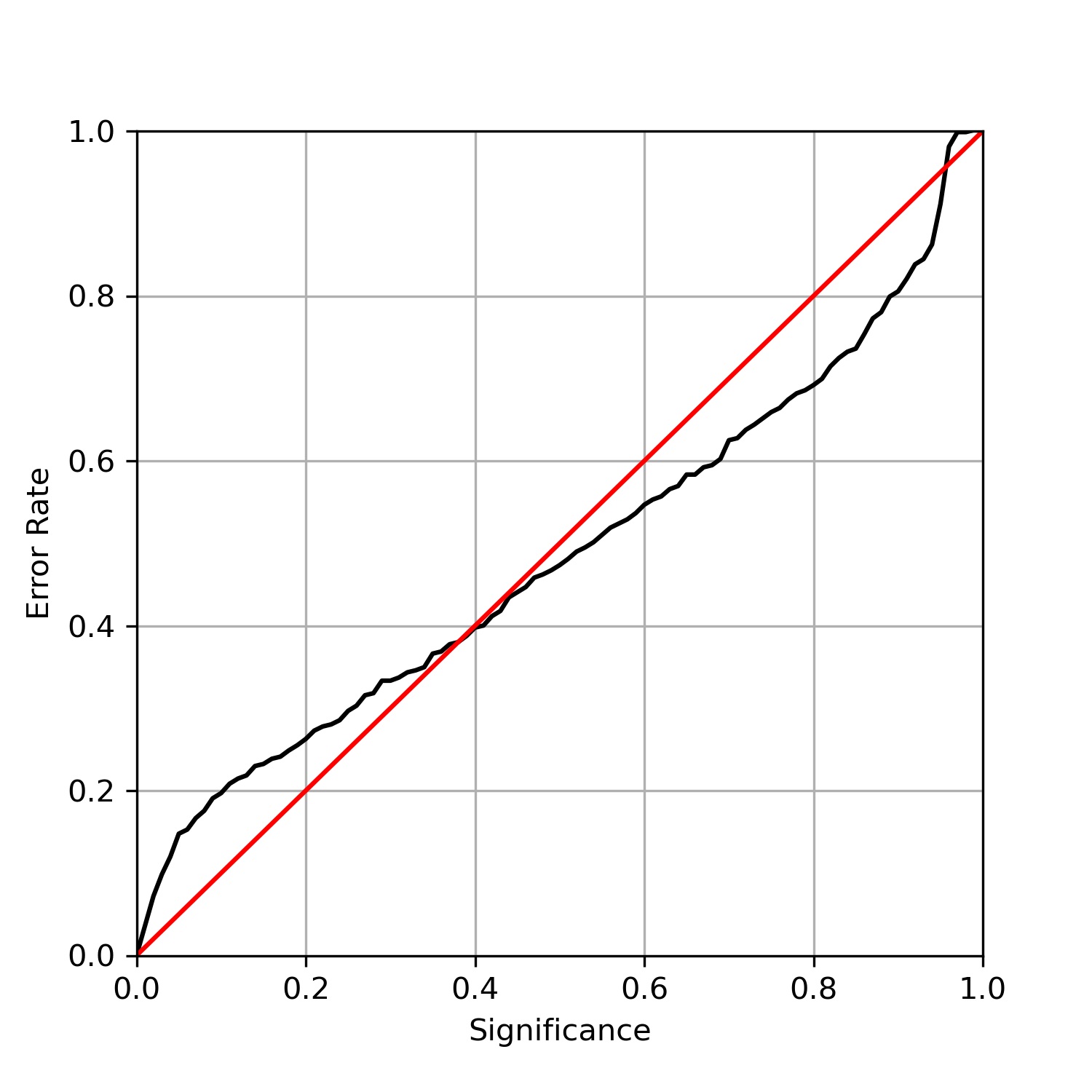}
        \caption{Baseline Condition}
        \label{fig: err-baselinecondition} 
    \end{subfigure}
    ~
    \begin{subfigure}[t]{0.45\textwidth}
        \centering
        \includegraphics[width=\textwidth]{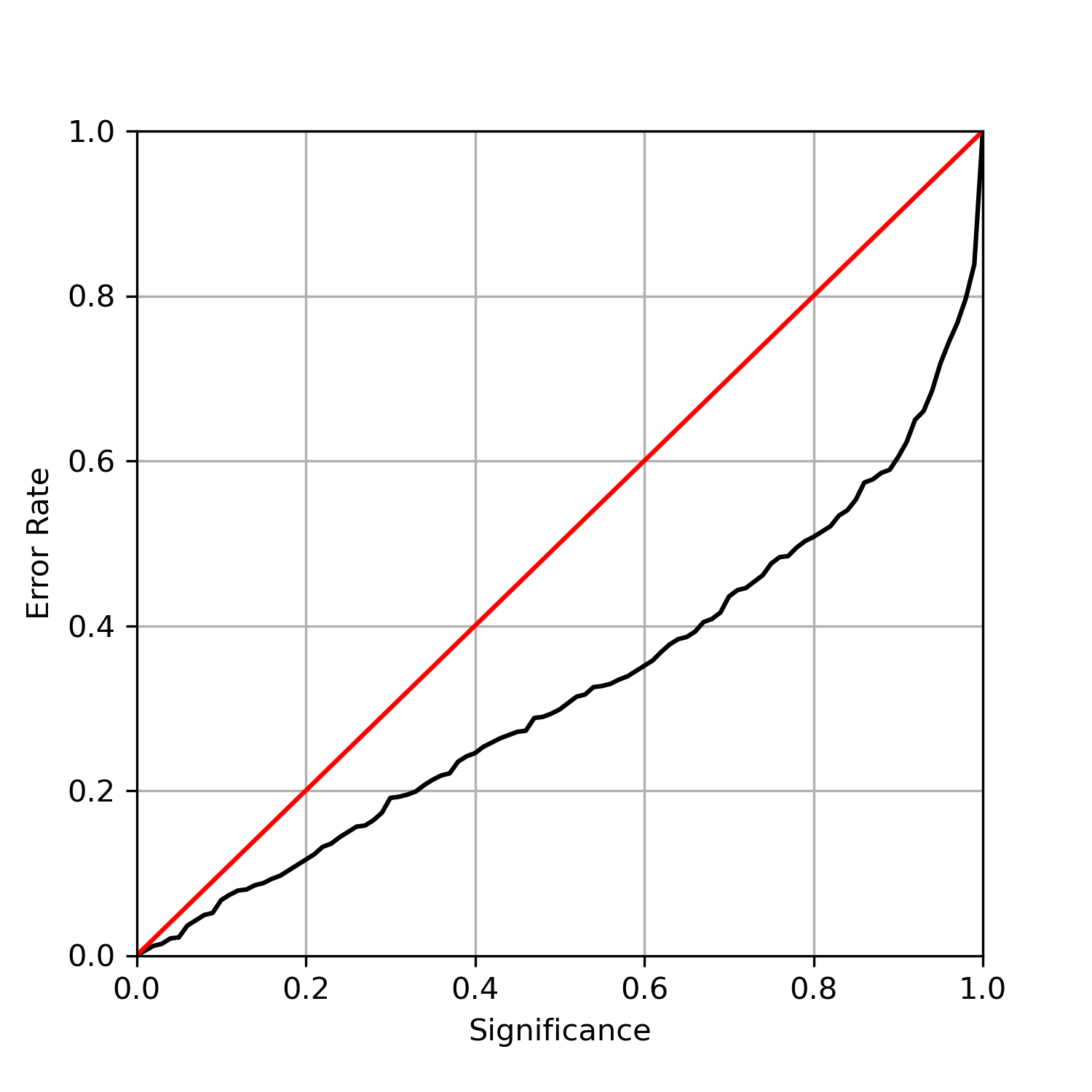}
        \caption{High Workload Nominal Condition}
        \label{fig: err-nominalcondition} 
    \end{subfigure}
    ~
    \begin{subfigure}[t]{0.45\textwidth}
        \centering
        \includegraphics[width=\textwidth]{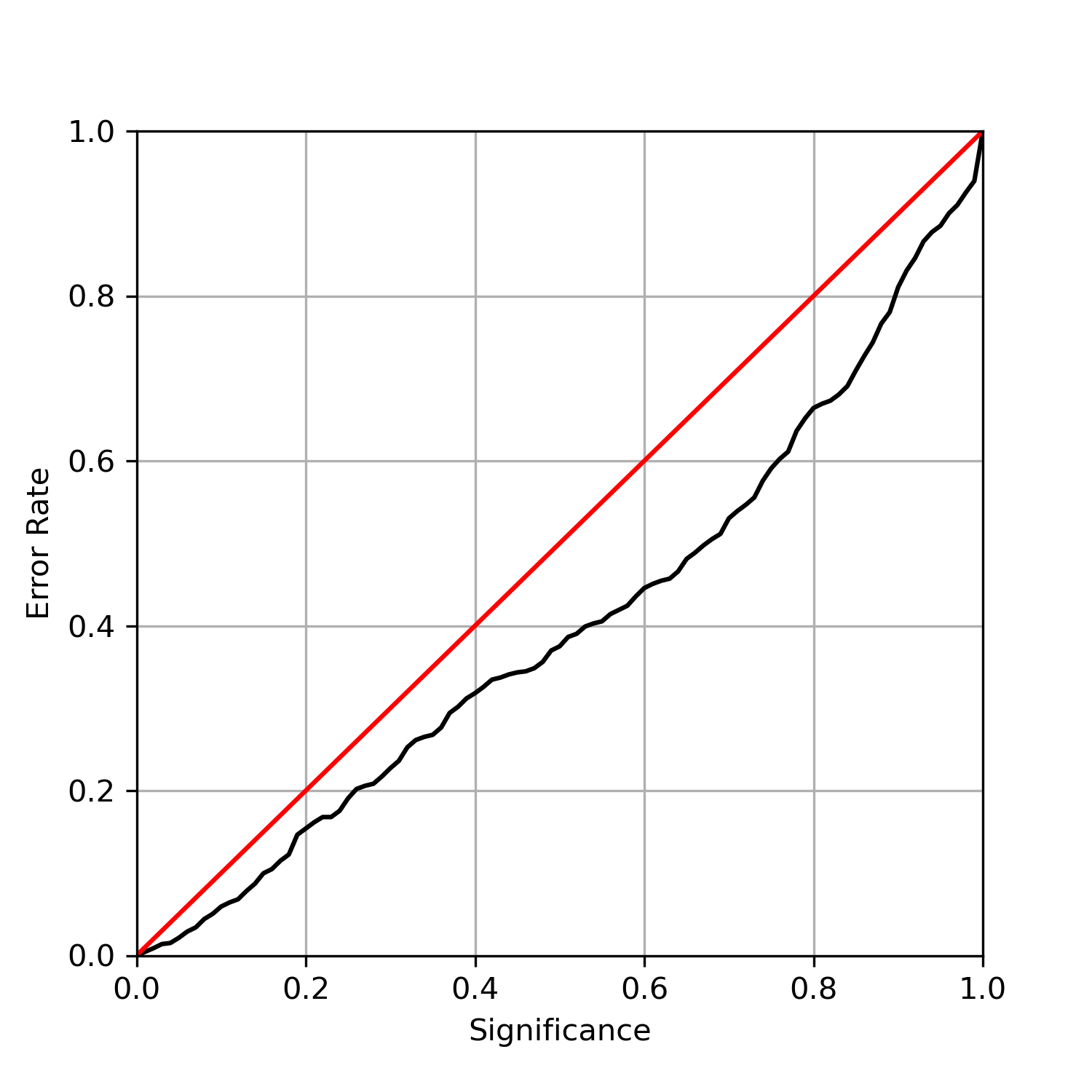}
        \caption{High Workload Off-Nominal Condition}
        \label{fig: err-offcondition} 
    \end{subfigure}
    \caption{The calibration plot illustrates the observed prediction error, which is the proportion of true labels that are not included in the prediction set, plotted against the pre-specified significance level $\varepsilon$, or the tolerated error rate. The conformal predictor is deemed valid only when the observed error rate is within the limit of $\varepsilon$, i.e., the observed error rate should align closely with the diagonal line representing the tolerated error rate for all significance levels. One of the key advantages of conformal predictors is their ability to offer valid predictions even when new examples are independently and identically distributed with the training examples. Additionally, we use Kolmogorov-Smirnov (K-S) test to test the distributions of predictions in the calibration set and test data under three conditions. In the K-S test, the null hypothesis is that the calibration and test samples are drawn from the same data distribution. The corresponding p-values obtained for three conditions are, (a) $1.98e-12$; (b) $3.68e-48$; (c) $2.18e-55$. All three values below $5\%$, are considered two-sided statistically significant. Thus, the hypothesis holds true.}
    \label{fig: err-vis}
\end{figure}

To further look at the coverages, we adopt another recently proposed figure to better show the prediction coverage violations \citep{olsson2022estimating}. In \Cref{fig: err-vis}, we show three figures for three simulation scenarios. The x-axis shows the tolerated error rate (the specified significant level), and the y-axis shows the fractions of failed prediction sets (the number of prediction samples where the ground truth label is not in the conformal prediction set). This property holds true in high workload scenarios but not in the baseline scenarios when the significance level is lower than $0.4$ or higher than $0.95$. 

\section{Conclusions\label{sec: conclusion}}
In this paper, we investigate the workload prediction problem. We formulate the problem into a time-series dynamic graph classification task with changing graph topologies. We demonstrate the effectiveness of this proposed method from real-world human-in-the-loop air traffic control simulations, in which participants are retired air traffic controllers. We show that traffic density features and traffic conflict features have a positive influence on workload predictions. Algorithm-wise, the graph-structured data-driven learning model outperforms the existing practices in workload prediction research literature (i.e., simple regressions, simple neural networks with handcrafted features).

\subsection{Limitations \label{subsec: limitations}}
There are several limitations. Firstly, we only have limited resources to conduct the HITL experiments in a simulation environment. Real-world scenarios can be immensely different from simulation scenarios, with either fewer or more deviations. Secondly, data quality is critical for developing a successful machine-learning algorithm. In this work, we have to use the corrected workload rating data due to the poor quality of the originally collected workload ratings, with only six retired ATCo participants \citep{lieber2020communications}. Another critical part is modeling the different ATC strategies adopted by different controllers, which can result in a multi-modal machine learning setup. The benefit of including ATCo strategies has also been discussed in the literature \citep{loft2007modeling}. Lastly, better algorithm development can help with improved workload prediction performance. The single prediction label made by either EGCU-O or EGCU-H can be improved, despite the data quality issue.

\subsection{Insights \label{subsec: insights}}
This work is beneficial for the non-intrusive, uninterrupted executive controller workload prediction, and it is purely based on the flight traffic data. First, we show that both traffic density and traffic conflict features contribute to higher prediction accuracy. Then, we show that model of the spatiotemporal airspace layout as a dynamic time-series graph learning problem has great potential for ATC workload level predictions. Additionally, we explore the possibility of further accuracy improvement by introducing a \textit{post-hoc} classification score processing process, namely conformal prediction, which can be used to generate multiple classification labels adaptively. 

Based on these insights, we propose several research directions that might be interesting to researchers,
\begin{itemize}
    \item We are expecting a significant performance improvement by conducting more HITL simulations or real-world ATC experiments, collecting additional high-quality data, and data-driven model refinement. Spatiotemporal graph learning is a popular theoretical research direction, and better graph learning model architecture is expected, which results in better workload prediction performances. 
    \item The window function setup for input-output data matching is flexible for any practical requirement in the real world. The length of the window determines the history length to be considered, while the stride size defines the prediction horizon.
    \item The workload prediction problem formulation can be alternated from workload rating classification to workload rating regression task. This setup is algorithm-wise more reasonable for uncertainty calibration with conformal prediction but requires significant experiment setup change. For instance, the current rating-based question prob in modified SWAT and NASA TLX will be modified to continuous variables. A considerable modification of the HITL simulations is desired.
    \item There are still several important questions that remain unanswered, such as how to incorporate real-time data and feedback into the prediction model and how to adapt the model to different types of air traffic control systems. Future research in this area could also explore the impact of other factors, such as weather conditions and aircraft type, on controller workload and safety.
    \item Several works of literature quires the validity of only using traffic-related factors to predict mental workload, where a significant part of pilot-controller interactions and feedback are missing \citep{loft2007modeling}. In further studies, we propose to build a predictive model that can consider reciprocal feedback interactions (i.e., the communication deviations \citep{djokic2010air} in \Cref{sec: human}) from a learning perspective. 
    \item Further studies can also combine trajectory prediction models such that the task demands can be predicted first and then perform workload forecast in real-time. In such a way, either deterministic or probabilistic trajectory prediction models can act as moderators of workload models \citep{corver2016predicting, pang2019recurrent, pang2021data, pang2022bayesian}. 
\end{itemize}

We believe the above discussions have practical implications for aviation authorities, airlines, and air traffic management providers. Specifically, our workload prediction model could be used to inform scheduling and staffing decisions, optimize resource allocation, and support proactive safety management. However, it is important to note that successfully implementing such interventions will require collaboration and communication across stakeholders.



\section*{Acknowledgment}
The research reported in this paper was supported by funds from NASA University Leadership Initiative program (Contract No. NNX17AJ86A, PI: Yongming Liu, Technical Officer: Anupa Bajwa). The support is gratefully acknowledged.

\bibliography{ref}

\begin{thebibliography}{96}
\expandafter\ifx\csname natexlab\endcsname\relax\def\natexlab#1{#1}\fi
\providecommand{\url}[1]{\texttt{#1}}
\providecommand{\href}[2]{#2}
\providecommand{\path}[1]{#1}
\providecommand{\DOIprefix}{doi:}
\providecommand{\ArXivprefix}{arXiv:}
\providecommand{\URLprefix}{URL: }
\providecommand{\Pubmedprefix}{pmid:}
\providecommand{\doi}[1]{\href{http://dx.doi.org/#1}{\path{#1}}}
\providecommand{\Pubmed}[1]{\href{pmid:#1}{\path{#1}}}
\providecommand{\bibinfo}[2]{#2}
\ifx\xfnm\relax \def\xfnm[#1]{\unskip,\space#1}\fi
\bibitem[{Abbass et~al.(2014)Abbass, Tang, Amin, Ellejmi and
  Kirby}]{abbass2014augmented}
\bibinfo{author}{Abbass, H.A.}, \bibinfo{author}{Tang, J.},
  \bibinfo{author}{Amin, R.}, \bibinfo{author}{Ellejmi, M.},
  \bibinfo{author}{Kirby, S.}, \bibinfo{year}{2014}.
\newblock \bibinfo{title}{Augmented cognition using real-time eeg-based
  adaptive strategies for air traffic control}, in:
  \bibinfo{booktitle}{Proceedings of the human factors and ergonomics society
  annual meeting}, \bibinfo{organization}{SAGE Publications Sage CA: Los
  Angeles, CA}. pp. \bibinfo{pages}{230--234}.
\bibitem[{Alvarsson et~al.(2021)Alvarsson, McShane, Norinder and
  Spjuth}]{alvarsson2021predicting}
\bibinfo{author}{Alvarsson, J.}, \bibinfo{author}{McShane, S.A.},
  \bibinfo{author}{Norinder, U.}, \bibinfo{author}{Spjuth, O.},
  \bibinfo{year}{2021}.
\newblock \bibinfo{title}{Predicting with confidence: using conformal
  prediction in drug discovery}.
\newblock \bibinfo{journal}{Journal of Pharmaceutical Sciences}
  \bibinfo{volume}{110}, \bibinfo{pages}{42--49}.
\bibitem[{Angelopoulos et~al.(2020)Angelopoulos, Bates, Malik and
  Jordan}]{angelopoulos2020uncertainty}
\bibinfo{author}{Angelopoulos, A.}, \bibinfo{author}{Bates, S.},
  \bibinfo{author}{Malik, J.}, \bibinfo{author}{Jordan, M.I.},
  \bibinfo{year}{2020}.
\newblock \bibinfo{title}{Uncertainty sets for image classifiers using
  conformal prediction}.
\newblock \bibinfo{journal}{arXiv preprint arXiv:2009.14193} .
\bibitem[{Angelopoulos and Bates(2021)}]{angelopoulos2021gentle}
\bibinfo{author}{Angelopoulos, A.N.}, \bibinfo{author}{Bates, S.},
  \bibinfo{year}{2021}.
\newblock \bibinfo{title}{A gentle introduction to conformal prediction and
  distribution-free uncertainty quantification}.
\newblock \bibinfo{journal}{arXiv preprint arXiv:2107.07511} .
\bibitem[{Angelopoulos et~al.(2022)Angelopoulos, Bates, Fisch, Lei and
  Schuster}]{angelopoulos2022conformal}
\bibinfo{author}{Angelopoulos, A.N.}, \bibinfo{author}{Bates, S.},
  \bibinfo{author}{Fisch, A.}, \bibinfo{author}{Lei, L.},
  \bibinfo{author}{Schuster, T.}, \bibinfo{year}{2022}.
\newblock \bibinfo{title}{Conformal risk control}.
\newblock \bibinfo{journal}{arXiv preprint arXiv:2208.02814} .
\bibitem[{Aric{\`o} et~al.(2016)Aric{\`o}, Borghini, Di~Flumeri, Colosimo,
  Bonelli, Golfetti, Pozzi, Imbert, Granger, Benhacene
  et~al.}]{arico2016adaptive}
\bibinfo{author}{Aric{\`o}, P.}, \bibinfo{author}{Borghini, G.},
  \bibinfo{author}{Di~Flumeri, G.}, \bibinfo{author}{Colosimo, A.},
  \bibinfo{author}{Bonelli, S.}, \bibinfo{author}{Golfetti, A.},
  \bibinfo{author}{Pozzi, S.}, \bibinfo{author}{Imbert, J.P.},
  \bibinfo{author}{Granger, G.}, \bibinfo{author}{Benhacene, R.}, et~al.,
  \bibinfo{year}{2016}.
\newblock \bibinfo{title}{Adaptive automation triggered by eeg-based mental
  workload index: a passive brain-computer interface application in realistic
  air traffic control environment}.
\newblock \bibinfo{journal}{Frontiers in human neuroscience}
  \bibinfo{volume}{10}, \bibinfo{pages}{539}.
\bibitem[{Balasubramanian et~al.(2014)Balasubramanian, Ho and
  Vovk}]{balasubramanian2014conformal}
\bibinfo{author}{Balasubramanian, V.}, \bibinfo{author}{Ho, S.S.},
  \bibinfo{author}{Vovk, V.}, \bibinfo{year}{2014}.
\newblock \bibinfo{title}{Conformal prediction for reliable machine learning:
  theory, adaptations and applications}.
\newblock \bibinfo{publisher}{Newnes}.
\bibitem[{Berghoff et~al.(2021)Berghoff, Neu and von
  Twickel}]{berghoff2021interplay}
\bibinfo{author}{Berghoff, C.}, \bibinfo{author}{Neu, M.}, \bibinfo{author}{von
  Twickel, A.}, \bibinfo{year}{2021}.
\newblock \bibinfo{title}{The interplay of ai and biometrics: Challenges and
  opportunities}.
\newblock \bibinfo{journal}{Computer} \bibinfo{volume}{54},
  \bibinfo{pages}{80--85}.
\bibitem[{Board.(2001)}]{national2001annual}
\bibinfo{author}{Board., N.T.S.}, \bibinfo{year}{2001}.
\newblock \bibinfo{title}{Annual review of aircraft accident data, us air
  carrier operations: Calendar year 2001 (rep. no. arc-06--01)}.
\bibitem[{Chatterji and Sridhar(2001)}]{chatterji2001measures}
\bibinfo{author}{Chatterji, G.}, \bibinfo{author}{Sridhar, B.},
  \bibinfo{year}{2001}.
\newblock \bibinfo{title}{Measures for air traffic controller workload
  prediction}, in: \bibinfo{booktitle}{1st AIAA, aircraft, technology
  Integration, and operations Forum}, p. \bibinfo{pages}{5242}.
\bibitem[{Chatterji and Sridhar(1999)}]{chatterji1999neural}
\bibinfo{author}{Chatterji, G.B.}, \bibinfo{author}{Sridhar, B.},
  \bibinfo{year}{1999}.
\newblock \bibinfo{title}{Neural network based air traffic controller workload
  prediction}, in: \bibinfo{booktitle}{Proceedings of the 1999 American Control
  Conference (Cat. No. 99CH36251)}, \bibinfo{organization}{IEEE}. pp.
  \bibinfo{pages}{2620--2624}.
\bibitem[{Cooke et~al.(2017)Cooke, Gorman and Kiekel}]{cooke2017communication}
\bibinfo{author}{Cooke, N.J.}, \bibinfo{author}{Gorman, J.C.},
  \bibinfo{author}{Kiekel, P.A.}, \bibinfo{year}{2017}.
\newblock \bibinfo{title}{Communication as team-level cognitive processing},
  in: \bibinfo{booktitle}{Macrocognition in teams}. \bibinfo{publisher}{CRC
  Press}, pp. \bibinfo{pages}{51--64}.
\bibitem[{Corver et~al.(2016)Corver, Unger and Grote}]{corver2016predicting}
\bibinfo{author}{Corver, S.C.}, \bibinfo{author}{Unger, D.},
  \bibinfo{author}{Grote, G.}, \bibinfo{year}{2016}.
\newblock \bibinfo{title}{Predicting air traffic controller workload:
  trajectory uncertainty as the moderator of the indirect effect of traffic
  density on controller workload through traffic conflict}.
\newblock \bibinfo{journal}{Human factors} \bibinfo{volume}{58},
  \bibinfo{pages}{560--573}.
\bibitem[{Crump(1979)}]{crump1979review}
\bibinfo{author}{Crump, J.H.}, \bibinfo{year}{1979}.
\newblock \bibinfo{title}{Review of stress in air traffic control: Its
  measurement and effects.}
\newblock \bibinfo{journal}{Aviation, Space, and Environmental Medicine} .
\bibitem[{Crutchfield and Rosenberg(2007)}]{crutchfield2007predicting}
\bibinfo{author}{Crutchfield, J.}, \bibinfo{author}{Rosenberg, C.},
  \bibinfo{year}{2007}.
\newblock \bibinfo{title}{Predicting subjective workload ratings: A comparison
  and synthesis of operational and theoretical models}.
\newblock \bibinfo{type}{Technical Report}. FEDERAL AVIATION ADMINISTRATION
  OKLAHOMA CITY OK CIVIL AEROMEDICAL INST.
\bibitem[{Delahaye and Puechmorel(2000)}]{delahaye2000air}
\bibinfo{author}{Delahaye, D.}, \bibinfo{author}{Puechmorel, S.},
  \bibinfo{year}{2000}.
\newblock \bibinfo{title}{Air traffic complexity: Towards an intrinsic metric},
  in: \bibinfo{booktitle}{Proceeding of the 3rd USA/Europe Air Traffic
  Management R and D Seminar}.
\bibitem[{Dhief et~al.(2020)Dhief, Wang, Liang, Alam, Schultz and
  Delahaye}]{dhief2020predicting}
\bibinfo{author}{Dhief, I.}, \bibinfo{author}{Wang, Z.},
  \bibinfo{author}{Liang, M.}, \bibinfo{author}{Alam, S.},
  \bibinfo{author}{Schultz, M.}, \bibinfo{author}{Delahaye, D.},
  \bibinfo{year}{2020}.
\newblock \bibinfo{title}{Predicting aircraft landing time in extended-tma
  using machine learning methods}, in: \bibinfo{booktitle}{ICRAT 2020, 9th
  International Conference for Research in Air Transportation}.
\bibitem[{Di~Stasi et~al.(2010)Di~Stasi, Marchitto, Antol{\'\i}, Baccino and
  Ca{\~n}as}]{di2010approximation}
\bibinfo{author}{Di~Stasi, L.L.}, \bibinfo{author}{Marchitto, M.},
  \bibinfo{author}{Antol{\'\i}, A.}, \bibinfo{author}{Baccino, T.},
  \bibinfo{author}{Ca{\~n}as, J.J.}, \bibinfo{year}{2010}.
\newblock \bibinfo{title}{Approximation of on-line mental workload index in atc
  simulated multitasks}.
\newblock \bibinfo{journal}{Journal of Air Transport Management}
  \bibinfo{volume}{16}, \bibinfo{pages}{330--333}.
\bibitem[{Djokic et~al.(2010)Djokic, Lorenz and Fricke}]{djokic2010air}
\bibinfo{author}{Djokic, J.}, \bibinfo{author}{Lorenz, B.},
  \bibinfo{author}{Fricke, H.}, \bibinfo{year}{2010}.
\newblock \bibinfo{title}{Air traffic control complexity as workload driver}.
\newblock \bibinfo{journal}{Transportation research part C: emerging
  technologies} \bibinfo{volume}{18}, \bibinfo{pages}{930--936}.
\bibitem[{Durso and Alexander(2010)}]{durso2010managing}
\bibinfo{author}{Durso, F.T.}, \bibinfo{author}{Alexander, A.L.},
  \bibinfo{year}{2010}.
\newblock \bibinfo{title}{Managing workload, performance, and situation
  awareness in aviation systems}, in: \bibinfo{booktitle}{Human factors in
  aviation}. \bibinfo{publisher}{Elsevier}, pp. \bibinfo{pages}{217--247}.
\bibitem[{EASA(2021)}]{easaAIv1}
\bibinfo{author}{EASA}, \bibinfo{year}{2021}.
\newblock \bibinfo{title}{Easa concept paper: Artificial intelligence roadmap:
  A human-centric approach to ai in aviation}.
\newblock
  \bibinfo{note}{\url{https://www.easa.europa.eu/en/downloads/109668/en}
  Accessed: 3-20-2023}.
\bibitem[{EASA(2023)}]{easaAIv2}
\bibinfo{author}{EASA}, \bibinfo{year}{2023}.
\newblock \bibinfo{title}{Easa concept paper: First usable guidance for level
  1\&2 machine learning applications: A deliverable of the easa ai roadmap}.
\newblock
  \bibinfo{note}{\url{https://www.easa.europa.eu/en/downloads/137631/en}
  Accessed: 3-20-2023}.
\bibitem[{Eckmann et~al.(1995)Eckmann, Kamphorst, Ruelle
  et~al.}]{eckmann1995recurrence}
\bibinfo{author}{Eckmann, J.P.}, \bibinfo{author}{Kamphorst, S.O.},
  \bibinfo{author}{Ruelle, D.}, et~al., \bibinfo{year}{1995}.
\newblock \bibinfo{title}{Recurrence plots of dynamical systems}.
\newblock \bibinfo{journal}{World Scientific Series on Nonlinear Science Series
  A} \bibinfo{volume}{16}, \bibinfo{pages}{441--446}.
\bibitem[{Edwards et~al.(2017)Edwards, Martin, Bienert and
  Mercer}]{edwards2017relationship}
\bibinfo{author}{Edwards, T.}, \bibinfo{author}{Martin, L.},
  \bibinfo{author}{Bienert, N.}, \bibinfo{author}{Mercer, J.},
  \bibinfo{year}{2017}.
\newblock \bibinfo{title}{The relationship between workload and performance in
  air traffic control: exploring the influence of levels of automation and
  variation in task demand}, in: \bibinfo{booktitle}{Human Mental Workload:
  Models and Applications: First International Symposium, H-WORKLOAD 2017,
  Dublin, Ireland, June 28-30, 2017, Revised Selected Papers 1},
  \bibinfo{organization}{Springer}. pp. \bibinfo{pages}{120--139}.
\bibitem[{Edwards et~al.(2012)Edwards, Sharples, Wilson and
  Kirwan}]{edwards2012factor}
\bibinfo{author}{Edwards, T.}, \bibinfo{author}{Sharples, S.},
  \bibinfo{author}{Wilson, J.R.}, \bibinfo{author}{Kirwan, B.},
  \bibinfo{year}{2012}.
\newblock \bibinfo{title}{Factor interaction influences on human performance in
  air traffic control: The need for a multifactorial model}.
\newblock \bibinfo{journal}{Work} \bibinfo{volume}{41},
  \bibinfo{pages}{159--166}.
\bibitem[{{FAA}(2020)}]{faareport}
\bibinfo{author}{{FAA}}, \bibinfo{year}{2020}.
\newblock \bibinfo{title}{Air traffic by the numbers}.
\newblock
  \bibinfo{howpublished}{\url{https://www.faa.gov/air_traffic/by_the_numbers/media/Air_Traffic_by_the_Numbers_2020.pdf}}.
\bibitem[{Fraccone et~al.(2011)Fraccone, Volovoi, Col{\'o}n and
  Blake}]{fraccone2011novel}
\bibinfo{author}{Fraccone, G.C.}, \bibinfo{author}{Volovoi, V.},
  \bibinfo{author}{Col{\'o}n, A.E.}, \bibinfo{author}{Blake, M.},
  \bibinfo{year}{2011}.
\newblock \bibinfo{title}{Novel air traffic procedures: investigation of
  off-nominal scenarios and potential hazards}.
\newblock \bibinfo{journal}{Journal of Aircraft} \bibinfo{volume}{48},
  \bibinfo{pages}{127--140}.
\bibitem[{Gianazza(2010)}]{gianazza2010forecasting}
\bibinfo{author}{Gianazza, D.}, \bibinfo{year}{2010}.
\newblock \bibinfo{title}{Forecasting workload and airspace configuration with
  neural networks and tree search methods}.
\newblock \bibinfo{journal}{Artificial intelligence} \bibinfo{volume}{174},
  \bibinfo{pages}{530--549}.
\bibitem[{Gianazza(2017)}]{gianazza2017learning}
\bibinfo{author}{Gianazza, D.}, \bibinfo{year}{2017}.
\newblock \bibinfo{title}{Learning air traffic controller workload from past
  sector operations}, in: \bibinfo{booktitle}{ATM Seminar, 12th USA/Europe Air
  Traffic Management R\&D Seminar}.
\bibitem[{Gianazza and Guittet(2006)}]{gianazza2006selection}
\bibinfo{author}{Gianazza, D.}, \bibinfo{author}{Guittet, K.},
  \bibinfo{year}{2006}.
\newblock \bibinfo{title}{Selection and evaluation of air traffic complexity
  metrics}, in: \bibinfo{booktitle}{2006 ieee/aiaa 25TH Digital Avionics
  Systems Conference}, \bibinfo{organization}{IEEE}. pp.
  \bibinfo{pages}{1--12}.
\bibitem[{Gopher and Donchin(1986)}]{gopher1986workload}
\bibinfo{author}{Gopher, D.}, \bibinfo{author}{Donchin, E.},
  \bibinfo{year}{1986}.
\newblock \bibinfo{title}{Workload: An examination of the concept.} .
\bibitem[{Gorman et~al.(2003)Gorman, Foltz, Kiekel, Martin and
  Cooke}]{gorman2003evaluation}
\bibinfo{author}{Gorman, J.C.}, \bibinfo{author}{Foltz, P.W.},
  \bibinfo{author}{Kiekel, P.A.}, \bibinfo{author}{Martin, M.J.},
  \bibinfo{author}{Cooke, N.J.}, \bibinfo{year}{2003}.
\newblock \bibinfo{title}{Evaluation of latent semantic analysis-based measures
  of team communications content}, in: \bibinfo{booktitle}{Proceedings of the
  Human Factors and Ergonomics Society annual meeting},
  \bibinfo{organization}{Sage Publications Sage CA: Los Angeles, CA}. pp.
  \bibinfo{pages}{424--428}.
\bibitem[{Hah et~al.(2006)Hah, Willems and Phillips}]{hah2006effect}
\bibinfo{author}{Hah, S.}, \bibinfo{author}{Willems, B.},
  \bibinfo{author}{Phillips, R.}, \bibinfo{year}{2006}.
\newblock \bibinfo{title}{The effect of air traffic increase on controller
  workload}, in: \bibinfo{booktitle}{Proceedings of the Human Factors and
  Ergonomics Society Annual Meeting}, \bibinfo{organization}{SAGE Publications
  Sage CA: Los Angeles, CA}. pp. \bibinfo{pages}{50--54}.
\bibitem[{Hancock and Meshkati(1988)}]{hancock1988human}
\bibinfo{author}{Hancock, P.A.}, \bibinfo{author}{Meshkati, N.},
  \bibinfo{year}{1988}.
\newblock \bibinfo{title}{Human mental workload}.
\newblock \bibinfo{publisher}{North-Holland Amsterdam}.
\bibitem[{Hart(2006)}]{hart2006nasa}
\bibinfo{author}{Hart, S.G.}, \bibinfo{year}{2006}.
\newblock \bibinfo{title}{Nasa-task load index (nasa-tlx); 20 years later}, in:
  \bibinfo{booktitle}{Proceedings of the human factors and ergonomics society
  annual meeting}, \bibinfo{organization}{Sage publications Sage CA: Los
  Angeles, CA}. pp. \bibinfo{pages}{904--908}.
\bibitem[{Hart and Staveland(1988)}]{hart1988development}
\bibinfo{author}{Hart, S.G.}, \bibinfo{author}{Staveland, L.E.},
  \bibinfo{year}{1988}.
\newblock \bibinfo{title}{Development of nasa-tlx (task load index): Results of
  empirical and theoretical research}, in: \bibinfo{booktitle}{Advances in
  psychology}. \bibinfo{publisher}{Elsevier}. volume~\bibinfo{volume}{52}, pp.
  \bibinfo{pages}{139--183}.
\bibitem[{Heng et~al.(2022)Heng, Wu, Wen et~al.}]{heng2022identifying}
\bibinfo{author}{Heng, Y.}, \bibinfo{author}{Wu, M.}, \bibinfo{author}{Wen,
  X.}, et~al., \bibinfo{year}{2022}.
\newblock \bibinfo{title}{Identifying key risk factors in air traffic
  controller workload by seir model}.
\newblock \bibinfo{journal}{Mathematical Problems in Engineering}
  \bibinfo{volume}{2022}.
\bibitem[{Hilburn(2004)}]{hilburn2004cognitive}
\bibinfo{author}{Hilburn, B.}, \bibinfo{year}{2004}.
\newblock \bibinfo{title}{Cognitive complexity in air traffic control: A
  literature review}.
\newblock \bibinfo{journal}{EEC note} \bibinfo{volume}{4},
  \bibinfo{pages}{1--80}.
\bibitem[{Hilburn and Flynn(2004)}]{hilburn2004toward}
\bibinfo{author}{Hilburn, B.}, \bibinfo{author}{Flynn, G.},
  \bibinfo{year}{2004}.
\newblock \bibinfo{title}{Toward a non-linear approach to modeling air traffic
  complexity}, in: \bibinfo{booktitle}{2nd Human Performance Situation
  Awareness and Automation Conference}.
\bibitem[{Histon et~al.(2002)Histon, Hansman, Aigoin, Delahaye and
  Puechmorel}]{histon2002introducing}
\bibinfo{author}{Histon, J.M.}, \bibinfo{author}{Hansman, R.J.},
  \bibinfo{author}{Aigoin, G.}, \bibinfo{author}{Delahaye, D.},
  \bibinfo{author}{Puechmorel, S.}, \bibinfo{year}{2002}.
\newblock \bibinfo{title}{Introducing structural considerations into complexity
  metrics}.
\newblock \bibinfo{journal}{Air Traffic Control Quarterly}
  \bibinfo{volume}{10}, \bibinfo{pages}{115--130}.
\bibitem[{Kallus et~al.(1999)Kallus, Van~Damme and
  Dittman}]{kallus1999integrated}
\bibinfo{author}{Kallus, K.}, \bibinfo{author}{Van~Damme, D.},
  \bibinfo{author}{Dittman, A.}, \bibinfo{year}{1999}.
\newblock \bibinfo{title}{Integrated job and task analysis of air traffic
  controllers: Phase 2}.
\newblock \bibinfo{journal}{Task analysis of en-route controllers (European Air
  Traffic Management Programme Rep. No. HUM. ET1. ST01. 1000-REP-04).
  EUROCONTROL, Brussels, Belgium} .
\bibitem[{Kantz(1994)}]{kantz1994quantifying}
\bibinfo{author}{Kantz, H.}, \bibinfo{year}{1994}.
\newblock \bibinfo{title}{Quantifying the closeness of fractal measures}.
\newblock \bibinfo{journal}{Physical Review E} \bibinfo{volume}{49},
  \bibinfo{pages}{5091}.
\bibitem[{Kipf and Welling(2016)}]{kipf2016semi}
\bibinfo{author}{Kipf, T.N.}, \bibinfo{author}{Welling, M.},
  \bibinfo{year}{2016}.
\newblock \bibinfo{title}{Semi-supervised classification with graph
  convolutional networks}.
\newblock \bibinfo{journal}{arXiv preprint arXiv:1609.02907} .
\bibitem[{Kirwan et~al.(2001)Kirwan, Scaife and
  Kennedy}]{kirwan2001investigating}
\bibinfo{author}{Kirwan, B.}, \bibinfo{author}{Scaife, R.},
  \bibinfo{author}{Kennedy, R.}, \bibinfo{year}{2001}.
\newblock \bibinfo{title}{Investigating complexity factors in uk air traffic
  management}.
\newblock \bibinfo{journal}{Human Factors and Aerospace Safety}
  \bibinfo{volume}{1}.
\bibitem[{Knorr and Walter(2011)}]{knorr2011trajectory}
\bibinfo{author}{Knorr, D.}, \bibinfo{author}{Walter, L.},
  \bibinfo{year}{2011}.
\newblock \bibinfo{title}{Trajectory uncertainty and the impact on sector
  complexity and workload}.
\newblock \bibinfo{journal}{SESAR Innovation Days} \bibinfo{volume}{29}.
\bibitem[{Koebbe and Mayer-Kress(1992)}]{koebbe1992use}
\bibinfo{author}{Koebbe, M.}, \bibinfo{author}{Mayer-Kress, G.},
  \bibinfo{year}{1992}.
\newblock \bibinfo{title}{Use of recurrence plots in the analysis of
  time-series data}, in: \bibinfo{booktitle}{SANTA FE INSTITUTE STUDIES IN THE
  SCIENCES OF COMPLEXITY-PROCEEDINGS VOLUME-},
  \bibinfo{organization}{Citeseer}. pp. \bibinfo{pages}{361--361}.
\bibitem[{Kopardekar and Magyarits(2003)}]{kopardekar2003measurement}
\bibinfo{author}{Kopardekar, P.}, \bibinfo{author}{Magyarits, S.},
  \bibinfo{year}{2003}.
\newblock \bibinfo{title}{Measurement and prediction of dynamic density}, in:
  \bibinfo{booktitle}{Proceedings of the 5th usa/europe air traffic management
  r \& d seminar}.
\bibitem[{Lei and Wasserman(2014)}]{lei2014distribution}
\bibinfo{author}{Lei, J.}, \bibinfo{author}{Wasserman, L.},
  \bibinfo{year}{2014}.
\newblock \bibinfo{title}{Distribution-free prediction bands for non-parametric
  regression}.
\newblock \bibinfo{journal}{Journal of the Royal Statistical Society: Series B:
  Statistical Methodology} , \bibinfo{pages}{71--96}.
\bibitem[{Li et~al.(2019a)Li, Lee, Chen and Khoo}]{li2019hybrid}
\bibinfo{author}{Li, F.}, \bibinfo{author}{Lee, C.H.}, \bibinfo{author}{Chen,
  C.H.}, \bibinfo{author}{Khoo, L.P.}, \bibinfo{year}{2019}a.
\newblock \bibinfo{title}{Hybrid data-driven vigilance model in traffic control
  center using eye-tracking data and context data}.
\newblock \bibinfo{journal}{Advanced Engineering Informatics}
  \bibinfo{volume}{42}, \bibinfo{pages}{100940}.
\bibitem[{Li et~al.(2019b)Li, Ying and Chuah}]{li2019grip}
\bibinfo{author}{Li, X.}, \bibinfo{author}{Ying, X.}, \bibinfo{author}{Chuah,
  M.C.}, \bibinfo{year}{2019}b.
\newblock \bibinfo{title}{Grip: Graph-based interaction-aware trajectory
  prediction}, in: \bibinfo{booktitle}{2019 IEEE Intelligent Transportation
  Systems Conference (ITSC)}, \bibinfo{organization}{IEEE}. pp.
  \bibinfo{pages}{3960--3966}.
\bibitem[{Liang et~al.(2020)Liang, Samtani, Guo and Yu}]{liang2020behavioral}
\bibinfo{author}{Liang, Y.}, \bibinfo{author}{Samtani, S.},
  \bibinfo{author}{Guo, B.}, \bibinfo{author}{Yu, Z.}, \bibinfo{year}{2020}.
\newblock \bibinfo{title}{Behavioral biometrics for continuous authentication
  in the internet-of-things era: An artificial intelligence perspective}.
\newblock \bibinfo{journal}{IEEE Internet of Things Journal}
  \bibinfo{volume}{7}, \bibinfo{pages}{9128--9143}.
\bibitem[{Lieber(2020)}]{lieber2020communications}
\bibinfo{author}{Lieber, C.}, \bibinfo{year}{2020}.
\newblock \bibinfo{title}{Communications Between Air Traffic Controllers and
  Pilots During Simulated Arrivals: Relation of Closed Loop Communication
  Deviations to Loss of Separation}.
\newblock Ph.D. thesis. Arizona State University.
\bibitem[{Lieber et~al.(2021)Lieber, Demir, Cooke and
  Ligda}]{lieber2021deviations}
\bibinfo{author}{Lieber, C.S.}, \bibinfo{author}{Demir, M.},
  \bibinfo{author}{Cooke, N.}, \bibinfo{author}{Ligda, S.},
  \bibinfo{year}{2021}.
\newblock \bibinfo{title}{Deviations in closed loop communications between air
  traffic controllers and pilots as a predictor of loss of separation}, in:
  \bibinfo{booktitle}{AIAA AVIATION 2021 FORUM}, p. \bibinfo{pages}{2320}.
\bibitem[{Ligda et~al.(2019)Ligda, Seeds, Harris, Lieber, Demir and
  Cooke}]{ligda2019monitoring}
\bibinfo{author}{Ligda, S.V.}, \bibinfo{author}{Seeds, M.L.},
  \bibinfo{author}{Harris, M.J.}, \bibinfo{author}{Lieber, C.S.},
  \bibinfo{author}{Demir, M.}, \bibinfo{author}{Cooke, N.},
  \bibinfo{year}{2019}.
\newblock \bibinfo{title}{Monitoring human performance in real-time for nas
  safety prognostics}, in: \bibinfo{booktitle}{AIAA Aviation 2019 Forum}, p.
  \bibinfo{pages}{3411}.
\bibitem[{Liu and Goebel(2018)}]{liu2018information}
\bibinfo{author}{Liu, Y.}, \bibinfo{author}{Goebel, K.}, \bibinfo{year}{2018}.
\newblock \bibinfo{title}{Information fusion for national airspace system
  prognostics: A nasa uli project}, in: \bibinfo{booktitle}{Proceedings of the
  10th Annual Conference of the Prognostics and Health Management Society, PHM,
  Philadelphia Center City, Philadelphia, PA, USA}, pp.
  \bibinfo{pages}{24--27}.
\bibitem[{Loft et~al.(2007)Loft, Sanderson, Neal and Mooij}]{loft2007modeling}
\bibinfo{author}{Loft, S.}, \bibinfo{author}{Sanderson, P.},
  \bibinfo{author}{Neal, A.}, \bibinfo{author}{Mooij, M.},
  \bibinfo{year}{2007}.
\newblock \bibinfo{title}{Modeling and predicting mental workload in en route
  air traffic control: Critical review and broader implications}.
\newblock \bibinfo{journal}{Human factors} \bibinfo{volume}{49},
  \bibinfo{pages}{376--399}.
\bibitem[{Lu et~al.(2022)Lu, Lemay, Chang, H{\"o}bel and
  Kalpathy-Cramer}]{lu2022fair}
\bibinfo{author}{Lu, C.}, \bibinfo{author}{Lemay, A.}, \bibinfo{author}{Chang,
  K.}, \bibinfo{author}{H{\"o}bel, K.}, \bibinfo{author}{Kalpathy-Cramer, J.},
  \bibinfo{year}{2022}.
\newblock \bibinfo{title}{Fair conformal predictors for applications in medical
  imaging}, in: \bibinfo{booktitle}{Proceedings of the AAAI Conference on
  Artificial Intelligence}, pp. \bibinfo{pages}{12008--12016}.
\bibitem[{Luo et~al.(2022)Luo, Zhao, Kuck, Ivanovic, Savarese, Schmerling and
  Pavone}]{luo2022sample}
\bibinfo{author}{Luo, R.}, \bibinfo{author}{Zhao, S.}, \bibinfo{author}{Kuck,
  J.}, \bibinfo{author}{Ivanovic, B.}, \bibinfo{author}{Savarese, S.},
  \bibinfo{author}{Schmerling, E.}, \bibinfo{author}{Pavone, M.},
  \bibinfo{year}{2022}.
\newblock \bibinfo{title}{Sample-efficient safety assurances using conformal
  prediction}, in: \bibinfo{booktitle}{Algorithmic Foundations of Robotics XV:
  Proceedings of the Fifteenth Workshop on the Algorithmic Foundations of
  Robotics}, \bibinfo{organization}{Springer}. pp. \bibinfo{pages}{149--169}.
\bibitem[{Majumdar and Ochieng(2002)}]{majumdar2002factors}
\bibinfo{author}{Majumdar, A.}, \bibinfo{author}{Ochieng, W.Y.},
  \bibinfo{year}{2002}.
\newblock \bibinfo{title}{Factors affecting air traffic controller workload:
  Multivariate analysis based on simulation modeling of controller workload}.
\newblock \bibinfo{journal}{Transportation Research Record}
  \bibinfo{volume}{1788}, \bibinfo{pages}{58--69}.
\bibitem[{Manning et~al.(2002)Manning, Mills, Fox, Pfleiderer and
  Mogilka}]{manning2002using}
\bibinfo{author}{Manning, C.A.}, \bibinfo{author}{Mills, S.H.},
  \bibinfo{author}{Fox, C.M.}, \bibinfo{author}{Pfleiderer, E.M.},
  \bibinfo{author}{Mogilka, H.J.}, \bibinfo{year}{2002}.
\newblock \bibinfo{title}{Using air traffic control taskload measures and
  communication events to predict subjective workload}.
\newblock \bibinfo{type}{Technical Report}. FEDERAL AVIATION ADMINISTRATION
  OKLAHOMA CITY OK CIVIL AEROMEDICAL INST.
\bibitem[{Marwan et~al.(2007a)Marwan, {Carmen Romano}, Thiel and
  Kurths}]{MARWAN2007237}
\bibinfo{author}{Marwan, N.}, \bibinfo{author}{{Carmen Romano}, M.},
  \bibinfo{author}{Thiel, M.}, \bibinfo{author}{Kurths, J.},
  \bibinfo{year}{2007}a.
\newblock \bibinfo{title}{Recurrence plots for the analysis of complex
  systems}.
\newblock \bibinfo{journal}{Physics Reports} \bibinfo{volume}{438},
  \bibinfo{pages}{237--329}.
\newblock \URLprefix
  \url{https://www.sciencedirect.com/science/article/pii/S0370157306004066},
  \DOIprefix\doi{https://doi.org/10.1016/j.physrep.2006.11.001}.
\bibitem[{Marwan et~al.(2007b)Marwan, Romano, Thiel and
  Kurths}]{marwan2007recurrence}
\bibinfo{author}{Marwan, N.}, \bibinfo{author}{Romano, M.C.},
  \bibinfo{author}{Thiel, M.}, \bibinfo{author}{Kurths, J.},
  \bibinfo{year}{2007}b.
\newblock \bibinfo{title}{Recurrence plots for the analysis of complex
  systems}.
\newblock \bibinfo{journal}{Physics reports} \bibinfo{volume}{438},
  \bibinfo{pages}{237--329}.
\bibitem[{Marwan et~al.(2002)Marwan, Wessel, Meyerfeldt, Schirdewan and
  Kurths}]{marwan2002recurrence}
\bibinfo{author}{Marwan, N.}, \bibinfo{author}{Wessel, N.},
  \bibinfo{author}{Meyerfeldt, U.}, \bibinfo{author}{Schirdewan, A.},
  \bibinfo{author}{Kurths, J.}, \bibinfo{year}{2002}.
\newblock \bibinfo{title}{Recurrence-plot-based measures of complexity and
  their application to heart-rate-variability data}.
\newblock \bibinfo{journal}{Physical review E} \bibinfo{volume}{66},
  \bibinfo{pages}{026702}.
\bibitem[{Masalonis et~al.(2003)Masalonis, Callaham and
  Wanke}]{masalonis2003dynamic}
\bibinfo{author}{Masalonis, A.J.}, \bibinfo{author}{Callaham, M.B.},
  \bibinfo{author}{Wanke, C.R.}, \bibinfo{year}{2003}.
\newblock \bibinfo{title}{Dynamic density and complexity metrics for realtime
  traffic flow management}, in: \bibinfo{booktitle}{Proceedings of the 5th
  USA/Europe Air Traffic Management R \& D Seminar},
  \bibinfo{organization}{Budapest, Hungary}. p. \bibinfo{pages}{139}.
\bibitem[{Mindlin and Gilmore(1992)}]{mindlin1992topological}
\bibinfo{author}{Mindlin, G.M.}, \bibinfo{author}{Gilmore, R.},
  \bibinfo{year}{1992}.
\newblock \bibinfo{title}{Topological analysis and synthesis of chaotic time
  series}.
\newblock \bibinfo{journal}{Physica D: Nonlinear Phenomena}
  \bibinfo{volume}{58}, \bibinfo{pages}{229--242}.
\bibitem[{Mogford et~al.(1995)Mogford, Guttman, Morrow and
  Kopardekar}]{mogford1995complexity}
\bibinfo{author}{Mogford, R.H.}, \bibinfo{author}{Guttman, J.},
  \bibinfo{author}{Morrow, S.}, \bibinfo{author}{Kopardekar, P.},
  \bibinfo{year}{1995}.
\newblock \bibinfo{title}{The complexity construct in air traffic control: A
  review and synthesis of the literature.} .
\bibitem[{Mohamed et~al.(2020)Mohamed, Qian, Elhoseiny and
  Claudel}]{mohamed2020social}
\bibinfo{author}{Mohamed, A.}, \bibinfo{author}{Qian, K.},
  \bibinfo{author}{Elhoseiny, M.}, \bibinfo{author}{Claudel, C.},
  \bibinfo{year}{2020}.
\newblock \bibinfo{title}{Social-stgcnn: A social spatio-temporal graph
  convolutional neural network for human trajectory prediction}, in:
  \bibinfo{booktitle}{Proceedings of the IEEE/CVF Conference on Computer Vision
  and Pattern Recognition}, pp. \bibinfo{pages}{14424--14432}.
\bibitem[{Nachreiner et~al.(2006)Nachreiner, Nickel and
  Meyer}]{nachreiner2006human}
\bibinfo{author}{Nachreiner, F.}, \bibinfo{author}{Nickel, P.},
  \bibinfo{author}{Meyer, I.}, \bibinfo{year}{2006}.
\newblock \bibinfo{title}{Human factors in process control systems: The design
  of human--machine interfaces}.
\newblock \bibinfo{journal}{Safety Science} \bibinfo{volume}{44},
  \bibinfo{pages}{5--26}.
\bibitem[{Olsson et~al.(2022)Olsson, Kartasalo, Mulliqi, Capuccini, Ruusuvuori,
  Samaratunga, Delahunt, Lindskog, Janssen, Blilie
  et~al.}]{olsson2022estimating}
\bibinfo{author}{Olsson, H.}, \bibinfo{author}{Kartasalo, K.},
  \bibinfo{author}{Mulliqi, N.}, \bibinfo{author}{Capuccini, M.},
  \bibinfo{author}{Ruusuvuori, P.}, \bibinfo{author}{Samaratunga, H.},
  \bibinfo{author}{Delahunt, B.}, \bibinfo{author}{Lindskog, C.},
  \bibinfo{author}{Janssen, E.A.}, \bibinfo{author}{Blilie, A.}, et~al.,
  \bibinfo{year}{2022}.
\newblock \bibinfo{title}{Estimating diagnostic uncertainty in artificial
  intelligence assisted pathology using conformal prediction}.
\newblock \bibinfo{journal}{Nature communications} \bibinfo{volume}{13},
  \bibinfo{pages}{7761}.
\bibitem[{Pang et~al.(2019)Pang, Yao, Hu and Liu}]{pang2019recurrent}
\bibinfo{author}{Pang, Y.}, \bibinfo{author}{Yao, H.}, \bibinfo{author}{Hu,
  J.}, \bibinfo{author}{Liu, Y.}, \bibinfo{year}{2019}.
\newblock \bibinfo{title}{A recurrent neural network approach for aircraft
  trajectory prediction with weather features from sherlock}, in:
  \bibinfo{booktitle}{AIAA Aviation 2019 Forum}, p. \bibinfo{pages}{3413}.
\bibitem[{Pang et~al.(2022)Pang, Zhao, Hu, Yan and Liu}]{pang2022bayesian}
\bibinfo{author}{Pang, Y.}, \bibinfo{author}{Zhao, X.}, \bibinfo{author}{Hu,
  J.}, \bibinfo{author}{Yan, H.}, \bibinfo{author}{Liu, Y.},
  \bibinfo{year}{2022}.
\newblock \bibinfo{title}{Bayesian spatio-temporal graph transformer network
  (b-star) for multi-aircraft trajectory prediction}.
\newblock \bibinfo{journal}{Knowledge-Based Systems} \bibinfo{volume}{249},
  \bibinfo{pages}{108998}.
\bibitem[{Pang et~al.(2021)Pang, Zhao, Yan and Liu}]{pang2021data}
\bibinfo{author}{Pang, Y.}, \bibinfo{author}{Zhao, X.}, \bibinfo{author}{Yan,
  H.}, \bibinfo{author}{Liu, Y.}, \bibinfo{year}{2021}.
\newblock \bibinfo{title}{Data-driven trajectory prediction with weather
  uncertainties: A bayesian deep learning approach}.
\newblock \bibinfo{journal}{Transportation Research Part C: Emerging
  Technologies} \bibinfo{volume}{130}, \bibinfo{pages}{103326}.
\bibitem[{Papadopoulos et~al.(2002)Papadopoulos, Proedrou, Vovk and
  Gammerman}]{papadopoulos2002inductive}
\bibinfo{author}{Papadopoulos, H.}, \bibinfo{author}{Proedrou, K.},
  \bibinfo{author}{Vovk, V.}, \bibinfo{author}{Gammerman, A.},
  \bibinfo{year}{2002}.
\newblock \bibinfo{title}{Inductive confidence machines for regression}, in:
  \bibinfo{booktitle}{Machine Learning: ECML 2002: 13th European Conference on
  Machine Learning Helsinki, Finland, August 19--23, 2002 Proceedings 13},
  \bibinfo{organization}{Springer}. pp. \bibinfo{pages}{345--356}.
\bibitem[{Pareja et~al.(2020)Pareja, Domeniconi, Chen, Ma, Suzumura, Kanezashi,
  Kaler, Schardl and Leiserson}]{pareja2020evolvegcn}
\bibinfo{author}{Pareja, A.}, \bibinfo{author}{Domeniconi, G.},
  \bibinfo{author}{Chen, J.}, \bibinfo{author}{Ma, T.},
  \bibinfo{author}{Suzumura, T.}, \bibinfo{author}{Kanezashi, H.},
  \bibinfo{author}{Kaler, T.}, \bibinfo{author}{Schardl, T.},
  \bibinfo{author}{Leiserson, C.}, \bibinfo{year}{2020}.
\newblock \bibinfo{title}{Evolvegcn: Evolving graph convolutional networks for
  dynamic graphs}, in: \bibinfo{booktitle}{Proceedings of the AAAI conference
  on artificial intelligence}, pp. \bibinfo{pages}{5363--5370}.
\bibitem[{Pham et~al.(2020)Pham, Alam and Duong}]{pham2020air}
\bibinfo{author}{Pham, D.T.}, \bibinfo{author}{Alam, S.},
  \bibinfo{author}{Duong, V.}, \bibinfo{year}{2020}.
\newblock \bibinfo{title}{An air traffic controller action
  extraction-prediction model using machine learning approach}.
\newblock \bibinfo{journal}{Complexity} \bibinfo{volume}{2020},
  \bibinfo{pages}{1--19}.
\bibitem[{Regulations(2017)}]{regulations2017title}
\bibinfo{author}{Regulations, F.A.}, \bibinfo{year}{2017}.
\newblock \bibinfo{title}{Title 14-aeronautics and space}.
\newblock \bibinfo{journal}{US government Printing Office, Washington, USA} .
\bibitem[{Reid and Nygren(1988)}]{reid1988subjective}
\bibinfo{author}{Reid, G.B.}, \bibinfo{author}{Nygren, T.E.},
  \bibinfo{year}{1988}.
\newblock \bibinfo{title}{The subjective workload assessment technique: A
  scaling procedure for measuring mental workload}, in:
  \bibinfo{booktitle}{Advances in psychology}. \bibinfo{publisher}{Elsevier}.
  volume~\bibinfo{volume}{52}, pp. \bibinfo{pages}{185--218}.
\bibitem[{Romano et~al.(2020)Romano, Sesia and
  Candes}]{romano2020classification}
\bibinfo{author}{Romano, Y.}, \bibinfo{author}{Sesia, M.},
  \bibinfo{author}{Candes, E.}, \bibinfo{year}{2020}.
\newblock \bibinfo{title}{Classification with valid and adaptive coverage}.
\newblock \bibinfo{journal}{Advances in Neural Information Processing Systems}
  \bibinfo{volume}{33}, \bibinfo{pages}{3581--3591}.
\bibitem[{Rose et~al.(1978)Rose, Jenkins, Hurst et~al.}]{rose1978air}
\bibinfo{author}{Rose, R.M.}, \bibinfo{author}{Jenkins, C.D.},
  \bibinfo{author}{Hurst, M.W.}, et~al., \bibinfo{year}{1978}.
\newblock \bibinfo{title}{Air traffic controller health change study: a
  prospective investigation of physical, psychological and work-related
  changes.}
\newblock \bibinfo{type}{Technical Report}. Civil Aerospace Medical Institute.
\bibitem[{Sadinle et~al.(2019)Sadinle, Lei and Wasserman}]{sadinle2019least}
\bibinfo{author}{Sadinle, M.}, \bibinfo{author}{Lei, J.},
  \bibinfo{author}{Wasserman, L.}, \bibinfo{year}{2019}.
\newblock \bibinfo{title}{Least ambiguous set-valued classifiers with bounded
  error levels}.
\newblock \bibinfo{journal}{Journal of the American Statistical Association}
  \bibinfo{volume}{114}, \bibinfo{pages}{223--234}.
\bibitem[{Salas and Fiore(2004)}]{salas2004team}
\bibinfo{author}{Salas, E.E.}, \bibinfo{author}{Fiore, S.M.},
  \bibinfo{year}{2004}.
\newblock \bibinfo{title}{Team cognition: Understanding the factors that drive
  process and performance.}
\newblock \bibinfo{publisher}{American Psychological Association}.
\bibitem[{Sharma et~al.(2022)Sharma, Iyer and Pant}]{sharma2022cognitive}
\bibinfo{author}{Sharma, K.}, \bibinfo{author}{Iyer, H.},
  \bibinfo{author}{Pant, R.}, \bibinfo{year}{2022}.
\newblock \bibinfo{title}{Cognitive ability criterion for expertise in air
  traffic control task}, in: \bibinfo{booktitle}{AIAA SCITECH 2022 Forum}, p.
  \bibinfo{pages}{2449}.
\bibitem[{Sheridan et~al.(2002)Sheridan, Sheridan, Maschinenbauingenieur,
  Sheridan and Sheridan}]{sheridan2002humans}
\bibinfo{author}{Sheridan, T.B.}, \bibinfo{author}{Sheridan, T.B.},
  \bibinfo{author}{Maschinenbauingenieur, K.}, \bibinfo{author}{Sheridan,
  T.B.}, \bibinfo{author}{Sheridan, T.B.}, \bibinfo{year}{2002}.
\newblock \bibinfo{title}{Humans and automation: System design and research
  issues}. volume \bibinfo{volume}{280}.
\newblock \bibinfo{publisher}{Human Factors and Ergonomics Society Santa
  Monica, CA}.
\bibitem[{Sridhar et~al.(1998)Sridhar, Sheth and Grabbe}]{sridhar1998airspace}
\bibinfo{author}{Sridhar, B.}, \bibinfo{author}{Sheth, K.S.},
  \bibinfo{author}{Grabbe, S.}, \bibinfo{year}{1998}.
\newblock \bibinfo{title}{Airspace complexity and its application in air
  traffic management}, in: \bibinfo{booktitle}{2nd USA/Europe Air Traffic
  Management R\&D Seminar}, \bibinfo{organization}{Federal Aviation
  Administration Washington, DC}. pp. \bibinfo{pages}{1--6}.
\bibitem[{Taha and Hanbury(2015)}]{taha2015metrics}
\bibinfo{author}{Taha, A.A.}, \bibinfo{author}{Hanbury, A.},
  \bibinfo{year}{2015}.
\newblock \bibinfo{title}{Metrics for evaluating 3d medical image segmentation:
  analysis, selection, and tool}.
\newblock \bibinfo{journal}{BMC medical imaging} \bibinfo{volume}{15},
  \bibinfo{pages}{1--28}.
\bibitem[{Tobaruela et~al.(2014)Tobaruela, Schuster, Majumdar, Ochieng,
  Martinez and Hendrickx}]{tobaruela2014method}
\bibinfo{author}{Tobaruela, G.}, \bibinfo{author}{Schuster, W.},
  \bibinfo{author}{Majumdar, A.}, \bibinfo{author}{Ochieng, W.Y.},
  \bibinfo{author}{Martinez, L.}, \bibinfo{author}{Hendrickx, P.},
  \bibinfo{year}{2014}.
\newblock \bibinfo{title}{A method to estimate air traffic controller mental
  workload based on traffic clearances}.
\newblock \bibinfo{journal}{Journal of Air Transport Management}
  \bibinfo{volume}{39}, \bibinfo{pages}{59--71}.
\bibitem[{Trapsilawati et~al.(2020)Trapsilawati, Herliansyah, Nugraheni,
  Fatikasari and Tissamodie}]{trapsilawati2020eeg}
\bibinfo{author}{Trapsilawati, F.}, \bibinfo{author}{Herliansyah, M.K.},
  \bibinfo{author}{Nugraheni, A.S.A.N.S.}, \bibinfo{author}{Fatikasari, M.P.},
  \bibinfo{author}{Tissamodie, G.}, \bibinfo{year}{2020}.
\newblock \bibinfo{title}{Eeg-based analysis of air traffic conflict:
  Investigating controllers’ situation awareness, stress level and brain
  activity during conflict resolution}.
\newblock \bibinfo{journal}{The Journal of Navigation} \bibinfo{volume}{73},
  \bibinfo{pages}{678--696}.
\bibitem[{Vogt et~al.(2006)Vogt, Hagemann and Kastner}]{vogt2006impact}
\bibinfo{author}{Vogt, J.}, \bibinfo{author}{Hagemann, T.},
  \bibinfo{author}{Kastner, M.}, \bibinfo{year}{2006}.
\newblock \bibinfo{title}{The impact of workload on heart rate and blood
  pressure in en-route and tower air traffic control}.
\newblock \bibinfo{journal}{Journal of psychophysiology} \bibinfo{volume}{20},
  \bibinfo{pages}{297--314}.
\bibitem[{Vovk et~al.(1999)Vovk, Gammerman and Saunders}]{vovk1999machine}
\bibinfo{author}{Vovk, V.}, \bibinfo{author}{Gammerman, A.},
  \bibinfo{author}{Saunders, C.}, \bibinfo{year}{1999}.
\newblock \bibinfo{title}{Machine-learning applications of algorithmic
  randomness} .
\bibitem[{Vovk et~al.(2005)Vovk, Gammerman and Shafer}]{vovk2005algorithmic}
\bibinfo{author}{Vovk, V.}, \bibinfo{author}{Gammerman, A.},
  \bibinfo{author}{Shafer, G.}, \bibinfo{year}{2005}.
\newblock \bibinfo{title}{Algorithmic learning in a random world}.
  volume~\bibinfo{volume}{29}.
\newblock \bibinfo{publisher}{Springer}.
\bibitem[{Wang et~al.(2015)Wang, Gong and Wen}]{wang2015air}
\bibinfo{author}{Wang, H.}, \bibinfo{author}{Gong, D.}, \bibinfo{author}{Wen,
  R.}, \bibinfo{year}{2015}.
\newblock \bibinfo{title}{Air traffic controllers workload forecasting method
  based on neural network}, in: \bibinfo{booktitle}{The 27th Chinese control
  and decision conference (2015 CCDC)}, \bibinfo{organization}{IEEE}. pp.
  \bibinfo{pages}{2460--2463}.
\bibitem[{Webber~Jr and Zbilut(1994)}]{webber1994dynamical}
\bibinfo{author}{Webber~Jr, C.L.}, \bibinfo{author}{Zbilut, J.P.},
  \bibinfo{year}{1994}.
\newblock \bibinfo{title}{Dynamical assessment of physiological systems and
  states using recurrence plot strategies}.
\newblock \bibinfo{journal}{Journal of applied physiology}
  \bibinfo{volume}{76}, \bibinfo{pages}{965--973}.
\bibitem[{Wee et~al.(2019)Wee, Lye and Pinheiro}]{wee2019integrated}
\bibinfo{author}{Wee, H.J.}, \bibinfo{author}{Lye, S.W.},
  \bibinfo{author}{Pinheiro, J.P.}, \bibinfo{year}{2019}.
\newblock \bibinfo{title}{An integrated highly synchronous, high resolution,
  real time eye tracking system for dynamic flight movement}.
\newblock \bibinfo{journal}{Advanced Engineering Informatics}
  \bibinfo{volume}{41}, \bibinfo{pages}{100919}.
\bibitem[{Wickens et~al.(2009)Wickens, Hooey, Gore, Sebok and
  Koenicke}]{wickens2009identifying}
\bibinfo{author}{Wickens, C.D.}, \bibinfo{author}{Hooey, B.L.},
  \bibinfo{author}{Gore, B.F.}, \bibinfo{author}{Sebok, A.},
  \bibinfo{author}{Koenicke, C.S.}, \bibinfo{year}{2009}.
\newblock \bibinfo{title}{Identifying black swans in nextgen: Predicting human
  performance in off-nominal conditions}.
\newblock \bibinfo{journal}{Human Factors} \bibinfo{volume}{51},
  \bibinfo{pages}{638--651}.
\bibitem[{Xiong et~al.(2023)Xiong, Wang, Tang, Cooke, Ligda, Lieber and
  Liu}]{xiong2023predicting}
\bibinfo{author}{Xiong, R.}, \bibinfo{author}{Wang, Y.}, \bibinfo{author}{Tang,
  P.}, \bibinfo{author}{Cooke, N.J.}, \bibinfo{author}{Ligda, S.V.},
  \bibinfo{author}{Lieber, C.S.}, \bibinfo{author}{Liu, Y.},
  \bibinfo{year}{2023}.
\newblock \bibinfo{title}{Predicting separation errors of air traffic
  controllers through integrated sequence analysis of multimodal behaviour
  indicators}.
\newblock \bibinfo{journal}{Advanced Engineering Informatics}
  \bibinfo{volume}{55}, \bibinfo{pages}{101894}.
\bibitem[{Zbilut and Webber~Jr(1992)}]{zbilut1992embeddings}
\bibinfo{author}{Zbilut, J.P.}, \bibinfo{author}{Webber~Jr, C.L.},
  \bibinfo{year}{1992}.
\newblock \bibinfo{title}{Embeddings and delays as derived from quantification
  of recurrence plots}.
\newblock \bibinfo{journal}{Physics letters A} \bibinfo{volume}{171},
  \bibinfo{pages}{199--203}.

\end{thebibliography}

\end{document}